\documentclass[10pt,conference]{IEEEtran}
\usepackage{cite}
\usepackage{graphicx}
\usepackage{balance}
\usepackage{epsfig}
\usepackage{epstopdf}
\usepackage{enumitem}
\usepackage{wrapfig}
\usepackage{amsmath}
\usepackage{amsthm}
\usepackage{amssymb}
\usepackage{mathrsfs}

\usepackage{algorithm}
\usepackage[]{algpseudocode}
\usepackage{bbm}
\usepackage{array}
\usepackage{tabularx}

\newcolumntype{Y}{>{\raggedright\arraybackslash}X}
\newcolumntype{L}[1]{>{\raggedright\arraybackslash}p{#1}}
\usepackage{amsfonts}

\usepackage{xcolor}
\ifCLASSINFOpdf
\else
\fi
\usepackage{colortbl}

\definecolor{darkblue}{rgb}{0.0, 0.0, 0.55}
\newcommand{\red}[1]{{\color{red}#1}}
\newcommand{\blue}[1]{{\color{blue}#1}}

\usepackage{lipsum}
\ifCLASSOPTIONcompsoc
    \usepackage[caption=false, font=normalsize, labelfont=sf, textfont=sf]{subfig}
\else
\usepackage[caption=false, font=footnotesize]{subfig}
\fi

\usepackage{soul}
\usepackage{multirow}
\usepackage{booktabs}
\usepackage{tabularx}
\usepackage{tikz}
\usepackage{tikzpagenodes}
\usetikzlibrary{calc}

\AddToHook{shipout/foreground}{%
	\ifnum\value{page}=1
	\begin{tikzpicture}[remember picture, overlay]
		\node[
		draw,
		text width=0.95\textwidth,
		align=left,
		font=\footnotesize,
		inner sep=5pt
		]
		at ($(current page header area.center) + (0,-5pt)$)
		{%
			This paper has been accepted for publication in the proceedings
			of  the IEEE International Conference on Mobile Ad-Hoc and Smart Systems
			(MASS), 2026. This is the authors' accepted version
			of the article. The final version published by IEEE is: A.~Alchaab, A.~Younis and D.~Pompili, ``SlicePlanner: Logic-Guided Agentic AI for Online
			QoS-Aware Network Slicing in Open RAN,'' in \textit{ Proc. of the IEEE International Conference
				on Mobile Ad-Hoc and Smart Systems (MASS)}, Hong Kong SAR, China, October 21–23, 2026.
		};
	\end{tikzpicture}%
	\fi
}

\begin{document}
\setlength{\columnsep}{0.17in}
\title
{SlicePlanner: Logic-Guided Agentic AI for Online QoS-Aware Network Slicing in Open RAN}
\author{\textbf{Adhwaa Al-Chaab†, Ayman Younis, Dario Pompili†}\\{†Department of Electrical and Computer Engineering}\\
{Rutgers University--New Brunswick, NJ, USA}\\
{E-mails: \{adhwaa.alchaab, a.younis, pompili\}@rutgers.edu
}\vspace{-.1in}}
\maketitle
\thispagestyle{empty}
\pagestyle{empty}

\maketitle
\thispagestyle{empty}
\pagestyle{empty}
\begin{abstract}
Adaptive network slicing in Open Radio Access Network (O-RAN) demands reconfiguration decisions that respond to live traffic conditions while remaining safe with respect to Service Level Agreement (SLA) constraints, a combination that purely data-driven controllers cannot reliably guarantee without continuous retraining. We present \textit{SlicePlanner}, a logic-guided closed-loop framework that treats each RAN reconfiguration as a controlled hypothesis: proposed from live and historical Key Performance Indicator (KPI) evidence, filtered through explicit SLA constraints and mined RAN control rules, and committed only when post-action measurements confirm improvement. \textit{SlicePlanner} comprises two cooperating agents: \textit{SliceSense}, which monitors live KPIs, retrieves context-conditioned evidence, and ranks candidate RAN actions through a surrogate-guided admissibility filter; and \textit{SliceGuard}, which validates post-action KPI responses, rolls back harmful configurations, and refreshes rules under distribution drift. Validated over the air using OpenAirInterface, FlexRIC, and commercial Sierra Wireless 5G devices, \textit{SlicePlanner} reduces SLA latency violations by up to $86\%$ over static slicing and $50\%$ over Deep Reinforcement Learning (DRL)-based slicing, while maintaining comparable cell throughput and auditable decisions without online retraining.
\end{abstract}

\section{Introduction}

\noindent\textbf{Overview.}
The evolution of mobile networks toward beyond fifth-generation (B5G) and sixth-generation (6G) architectures requires a fundamental shift in how network resources are managed~\cite{ahmadi2025towards}. Applications such as real-time video streaming, voice communication, industrial automation, and extended reality impose distinct and time-varying Quality of Service (QoS) requirements on the underlying network. Meeting these requirements across multiple users, applications, and channel conditions requires low-latency, fine-grained resource-allocation decisions conditioned on the near-real-time state of the network~\cite{bonati2023openran,younis2019demo}. Network slicing enables differentiated service delivery by partitioning a shared physical infrastructure into logically isolated virtual networks,
each configured for a specific service objective~\cite{wang2024intelligent}. A single sliced deployment can support throughput-intensive video traffic, latency-sensitive industrial services, and reliability-constrained communication flows. The
Open Radio Access Network (O-RAN) architecture makes this programmability operational by disaggregating the RAN into standardized components, including
the Radio Unit (RU), Distributed Unit (DU), and Central Unit (CU), and by
exposing network-state and control interfaces through the RAN Intelligent
Controller (RIC). These interfaces allow external control applications,
including xApps, to monitor Key Performance Indicators (KPIs) and enforce slice-aware configuration
decisions~\cite{polese2023understanding}.
\begin{figure}[t]
\centering
\includegraphics[width=0.95\columnwidth]{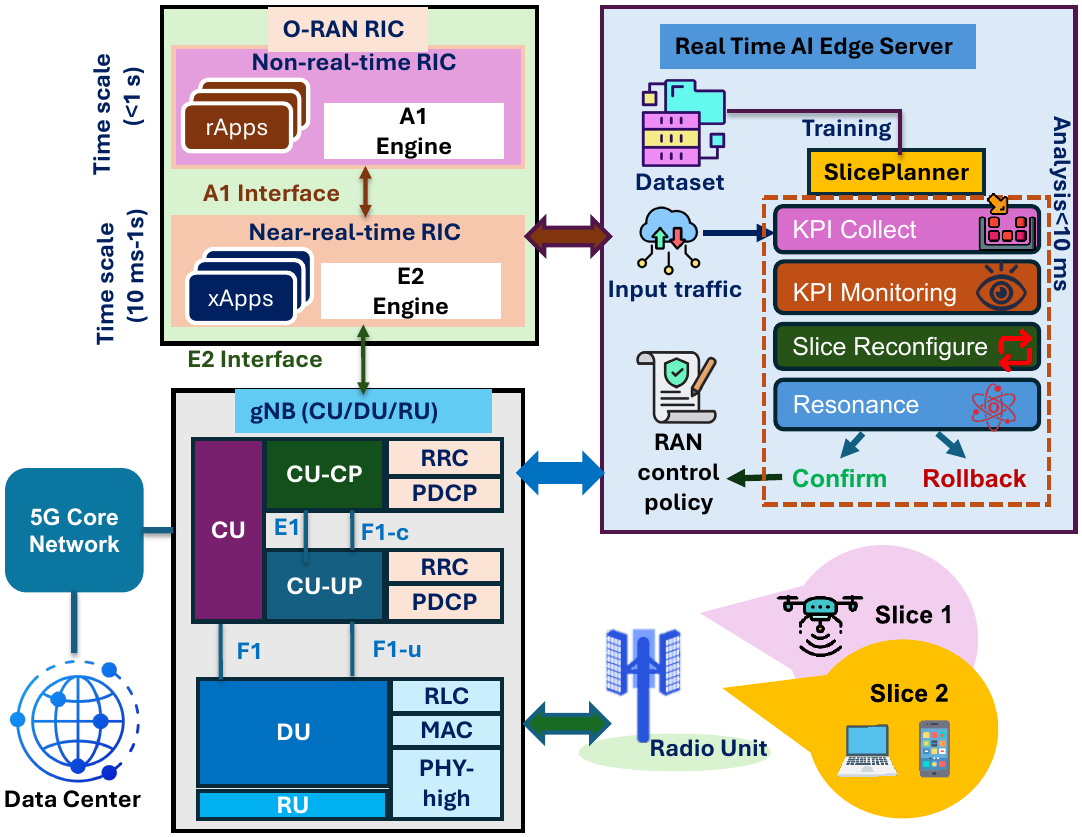}
\vspace{-0.06in}
\caption{SlicePlanner overview: closed-loop QoS-driven RAN reconfiguration with KPI monitoring, logic-guided selection, and post-action validation.}
\label{fig:NET}
\vspace{-0.18in}
\end{figure}

\noindent\textbf{Motivation.}
Adaptive network slicing in O-RAN requires decisions that are both
\emph{responsive} and \emph{safe}. Slice configurations need to adapt to
non-stationary traffic demand, User Equipment (UE) mobility, changing radio conditions, and
service-specific QoS requirements while preserving Service-Level Agreement (SLA) feasibility, slice
isolation, and configuration stability. This creates a fundamental systems-level challenge: the controller should react quickly to live network conditions
without selecting actions that improve one slice at the expense of another
or violate RAN operational constraints.

Existing slicing solutions address this challenge only partially.
Rule-based controllers are interpretable but often too rigid for dynamic
deployments~\cite{bonati2023openran}. For example, a threshold-based
controller that triggers Physical Resource Block (PRB) reallocation when packet loss exceeds a fixed
threshold cannot distinguish between a transient channel fade and a
persistent traffic overload~\cite{lacava2025dapps}.
Supervised-learning approaches depend on the coverage of the training data
and may degrade when the live network differs from the training
distribution~\cite{liu2021onslicing}. For instance, a classifier trained
on daytime traffic patterns may misclassify congestion events during
off-peak hours with atypical user distributions. Reinforcement
learning (RL) policies can optimize long-term objectives, but they
typically require extensive training, careful exploration, and retuning
when network behavior changes~\cite{alchaab2025slice,lotfi2025prompt}.
A Deep Deterministic Policy Gradient (DDPG)-based slicing agent trained under a fixed three-slice configuration,
for example, may require retraining or retuning when a new slice is
introduced or when traffic arrival rates shift outside the training
distribution~\cite{dai2024ran}.

A more fundamental limitation is that many data-driven slicing methods treat reconfiguration as an end-to-end state-to-action mapping: the controller
observes KPIs and directly outputs a resource allocation or RAN parameter update~\cite{liu2021onslicing,lotfi2025prompt}. This design can work well under familiar conditions but leaves critical networking structure implicit. SLA bounds, PRB-budget constraints, feasible parameter ranges, slice-coupling effects, and configuration stability are not merely statistical correlations; they
are operational constraints that should hold regardless of model confidence
or traffic distribution~\cite{dai2024ran, polese2023understanding}. As a result, a learned
policy may achieve high average reward during training yet still recommend an unsafe action under an unseen traffic mix, degraded channel state, or
shifted service priority. 
Recent neurosymbolic networking work motivates treating network knowledge
as explicit constraints that guide model outputs at decision time, rather than requiring the model to infer all constraints from data
alone~\cite{hakim2025neuro, gong2024zoom2net}.

\begin{figure}[t]
\centering
\includegraphics[width=.80\columnwidth,height=1.35in]{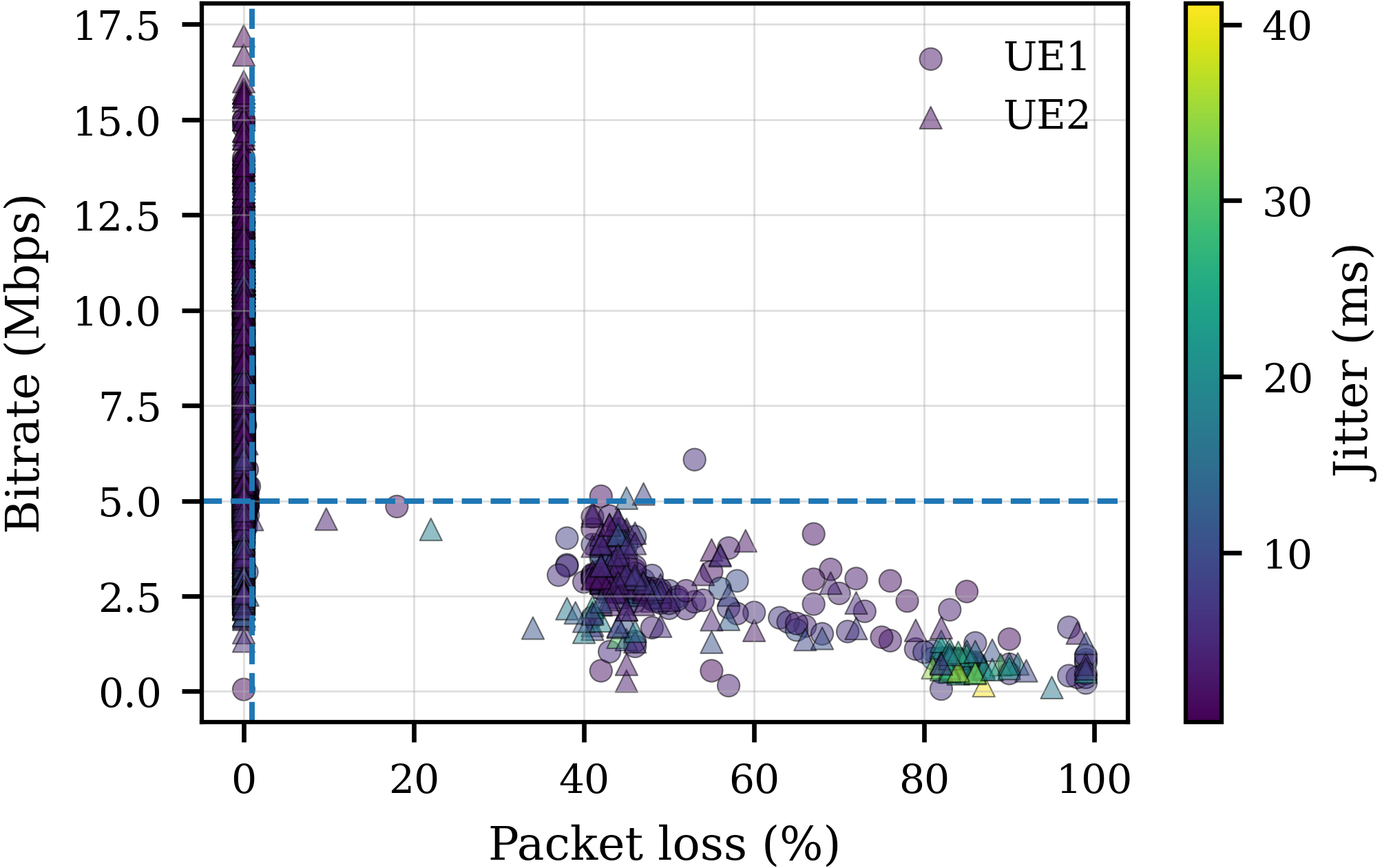}
\caption{QoS state space from over-the-air O-RAN testbed measurements. Points denote UE intervals; packet loss and bitrate define the QoS projection, while color shows jitter. Dashed lines mark the 5 Mbps bitrate target and 1\% loss threshold; violations trigger SlicePlanner reconfiguration.}
\label{fig:qos_state_space}
\vspace{-0.18in}
\end{figure}

Motivated by this principle, SlicePlanner formulates adaptive slicing as a logic-guided closed-loop control problem, as shown in Fig.~\ref{fig:NET}. Each reconfiguration is treated as a \emph{controlled hypothesis}: it is proposed using live and historical evidence, checked against network logic before execution, and accepted only when measured KPIs confirm its benefit. Fig.~\ref{fig:qos_state_space} supports this design using over-the-air O-RAN testbed measurements. The plotted UE states show that QoS degradation is multi-dimensional, appearing through excessive packet loss, insufficient bitrate, or jitter inflation. This motivates a multi-KPI degradation detector rather than a single-threshold monitor. The same QoS envelope is also used after reconfiguration: a candidate action is kept only if the affected UE or
slice moves toward the admissible region; otherwise, the previous stable configuration is restored.

\textbf{The main contributions of this work are as follows},
\begin{itemize}
    \item We propose SlicePlanner, a logic-guided closed-loop framework 
    for online QoS-aware network slicing in cloud-native O-RAN deployments. 
   \item We introduce an offline knowledge-construction procedure that mines RAN control rules from historical KPI records and operator-defined SLA conditions. These rules extend manual constraints with data-consistent relations among load, bandwidth, channel quality, and post-action KPI responses, and are automatically refreshed when live observations indicate distribution drift.

    \item We design the \emph{SliceSense Agent}, which performs live KPI monitoring, diagnoses degradation, ranks context-aware candidates via normalized gain, and verifies feasibility before reconfiguration.

    \item We design the \emph{SliceGuard Agent}, which measures post-action KPI  improvement, accepts beneficial updates, rolls back harmful or 
    ineffective configurations, and triggers rule refresh and controlled 
    re-exploration when the rule-inconsistency score exceeds a 
    defined threshold.
    \item We deploy SlicePlanner on a cloud-native 5G stack and validate it on an over-the-air O-RAN testbed using OpenAirInterface (OAI), FlexRIC, and commercial Sierra Wireless UEs. Experimental results show that SlicePlanner reduces SLA latency violations while maintaining comparable cell throughput across all evaluated scenarios.

\end{itemize}

\noindent \textbf{Paper Organization.} 
Sect.~\ref{sec:related_work} surveys related work. 
Sect.~\ref{sec:problem} presents the system model and problem formulation. 
Sect.~\ref{sec:proposed_work} details the SlicePlanner framework. 
Sect.~\ref{sec:performance_eval} evaluates performance via simulations 
and O-RAN testbed experiments. 
Sect.~\ref{sec:conclusion} concludes the paper and outlines future directions.
\section{Related Work}\label{sec:related_work}
Network slicing, O-RAN programmability, and AI-driven network automation have attracted considerable attention in both academia and industry in recent years.

\textbf{O-RAN Slicing and Resource Management.}
Network slicing has been extensively studied as a mechanism for providing differentiated QoS guarantees over shared physical infrastructure. In~\cite{salhab20215g}, the authors propose a machine learning framework for dynamic PRB allocation across network slices, demonstrating improved resource efficiency compared with static allocation baselines. The work in~\cite{polese2023understanding} provides a comprehensive survey of O-RAN architectures and identifies slice-level resource control as a key use case for near-real-time RIC applications. The work in~\cite{bonati2023openran} evaluates open-source RAN platforms for experimental O-RAN deployments and demonstrates the feasibility of closed-loop RAN control in real-world testbed environments. Although these works establish the infrastructure and control interfaces needed for programmable slicing, they leave open the problem of validation-aware online adaptation under changing network conditions.

\textbf{ML-Based RAN Control and Slice Optimization.}
Machine learning has been widely used to automate RAN configuration and slice allocation. Prior work includes online DRL for SLA-aware end-to-end slicing~\cite{liu2021onslicing}, dual-timescale slicing that separates \emph{long-term resource allocation} from \emph{Short-term resource scheduling}~\cite{alchaab2025slice}, prompt-tuned Large Language Model~(LLM) guidance for DRL-based O-RAN slicing~\cite{lotfi2025prompt}, and adaptive QoS-aware prioritized scheduling based on guaranteed allocation, urgency weighting, and priority-based round-robin control~\cite{yaqoob2026adaptive}. Although effective in their target settings, these approaches generally depend on fixed parameters learned from historical data. Under traffic, channel, or service shifts, performance may degrade until retraining is completed, limiting their use for near-real-time slice management.

\textbf{LLM and Agentic AI for Network Automation.}
LLMs are increasingly applied to network management and automation. Prior work includes LLM-driven scheduler generation with over-the-air testing in real 5G networks~\cite{elkael2025allstar}, zero-touch O-RAN deployment from natural-language intents~\cite{maxenti2026autoran}, hierarchical multi-agent control across timescales and protocol layers~\cite{elkael2025agentran}, trustworthy Artificial General Intelligence (AGI)-driven network management with symbiotic agents~\cite{chatzistefanidis2025symbiotic}, and structured LLM prompting for adaptive O-RAN resource allocation~\cite{wu2025llm}. These studies demonstrate the promise of LLM-assisted automation, but they do not couple online slicing decisions with live KPI validation, rollback, and retraining-free parameter adaptation.

Collectively, prior work establishes the feasibility of AI-driven RAN control, but mainly targets scheduling, power control, intent interpretation, or single-loop optimization. It does not directly address QoS-aware slice allocation across multiple RAN parameters with explicit validation and automatic rollback. \emph{SlicePlanner addresses this gap by transforming network slicing into a closed-loop, evidence-driven control process.} It integrates live O-RAN telemetry, an online KPI database, and structured reasoning to detect degradation, apply candidate slice reconfigurations, and validate their impact using post-action KPI feedback.

\section{Problem Formulation}
\label{sec:problem}

\subsection{System Model}
We consider a cloud-native disaggregated O-RAN network deployed on a container orchestration platform, such as Kubernetes. The network consists
of a single cell served by a gNodeB with a Central Unit (CU) and Distributed
Unit (DU) functional split. The system supports $N$ User Equipment (UE)
terminals, and each UE is associated with one of $S$ network slices. Each
slice $s \in \{1,\ldots,S\}$ serves a distinct application class with
differentiated QoS requirements, including enhanced Mobile Broadband
(eMBB), Ultra-Reliable and Low-Latency Communications (URLLC), and massive
Machine-Type Communications (mMTC). The RAN is controlled by the near-real-time RAN Intelligent Controller
(near-RT RIC), which operates at control timescales between 10\,ms and
1\,s. The near-RT RIC hosts a Radio Control Slice xApp (RC-Slice xApp) that applies slice-allocation and RAN-parameter control decisions, while RAN-side performance measurements are obtained through the Key Performance Measurement xApp (KPM xApp) over the O-RAN E2 interface. The system model is illustrated in Fig.~\ref{fig:NET}. At each control interval, the near-RT RIC observes RAN-side KPIs through
the E2 interface, including E2SM-KPM reports from the gNB, CU, and DU.
This separation distinguishes radio-side measurements reported through O-RAN E2
from service- and flow-level analytics obtained from the 5G core. 

Let $\mathbf{x}_t$ denote the observed network context at control interval
$t$. This context includes recent RAN-side KPIs, slice traffic demand, active UEs, and traffic statistics collected over a sliding observation window of $K$ intervals. Let $\bar{R}_{s}(t)$ and
$\bar{L}_{s}(t)$ denote the average downlink throughput and round-trip latency experienced by slice $s$ over this window, respectively. These per-slice KPIs are computed as,
\begin{equation}
\small
\bar{R}_{s}(t)=\frac{1}{K}\sum_{k=t-K+1}^{t}R_s(k),
\quad
\bar{L}_{s}(t)=\frac{1}{K}\sum_{k=t-K+1}^{t}L_s(k),
\label{eq:avg_kpis}
\end{equation}
where $R_{s}(k)$ and $L_{s}(k)$ are the instantaneous throughput and
latency measurements for slice $s$ at control interval $k$. Each slice is
associated with a minimum throughput requirement $R_{s}^{\min}$ and a
maximum latency requirement $L_{s}^{\max}$, which together define the slice SLA~\cite{alchaab2025slice}.
\subsection{SLA-Aware Network Slicing Optimization Problem}

Let $\mathcal{P}=\{p_1,p_2,\ldots,p_M\}$ denote the set of RAN parameters under management. In our experimental testbed, this set includes slicing parameters (PRB allocation, resource-share control) and radio parameters (uplink transmit power, downlink/uplink carrier bandwidth). Each parameter
$p_m \in \mathcal{P}$ takes values from a discrete admissible set, $\mathcal{V}_m =
\left\{
v_m^{(1)}, v_m^{(2)}, \ldots,
v_m^{(|\mathcal{V}_m|)}
\right\}$, defined by the network constraints. Let $v_m(t)$ denote the active value of parameter $p_m$ at time $t$, and let $\mathbf{v}(t) = [v_1(t),\ldots,v_M(t)]$ denote the current RAN configuration. In this work, SlicePlanner performs \emph{one-parameter candidate adjustment per control step}. This reduces the search space, limits unsafe exploration, and preserves operator interpretability. Given a parameter $p_m$ associated with the observed QoS degradation, the objective is to select a candidate value $v \in \mathcal{V}_m$ that improves aggregate slice-level QoS relative to the current value $v_0=v_m(t)$.
The historical KPI database stores context-conditioned observations of the
form, $(\mathbf{x}_t, v, \mathbf{y}_t)$, where $\mathbf{x}_t$ is the observed network context, $v$ is the applied
candidate value of parameter $p_m$, and $\mathbf{y}_t$ contains the
resulting slice-level KPIs. Based on this database, let
$\widehat{R}_{s}(v,\mathbf{x}_t)$ and
$\widehat{L}_{s}(v,\mathbf{x}_t)$ denote the estimated throughput and
latency of slice $s$ under candidate value $v$ and context $\mathbf{x}_t$.
The aggregate throughput and latency are defined as,
\begin{equation}
\small
\widehat{R}(v,\mathbf{x}_t)=\sum_{s=1}^{S}\alpha_s\widehat{R}_{s}(v,\mathbf{x}_t),
\quad
\widehat{L}(v,\mathbf{x}_t)=\sum_{s=1}^{S}\alpha_s\widehat{L}_{s}(v,\mathbf{x}_t).
\label{eq:agg_kpis}
\end{equation}
where $\alpha_s \geq 0$ denotes the priority weight of slice $s$, with
$\sum_{s=1}^{S}\alpha_s = 1$. The SLA-aware parameter selection problem is formulated as,
\begin{subequations}
\label{eq:opt}
\begin{align}
\max_{v \in \mathcal{V}_m} \quad
& G(v;\mathbf{x}_t)
= w_R \Delta_R(v;\mathbf{x}_t)
+ w_L \Delta_L(v;\mathbf{x}_t),
\label{eq:opt_a}
\\
\textrm{s.t.} \quad
& \widehat{R}_{s}^{-}(v,\mathbf{x}_t)
\geq R_{s}^{\min},
\quad \forall s \in \{1,\ldots,S\},
\label{eq:opt_b}
\\
& \widehat{L}_{s}^{+}(v,\mathbf{x}_t)
\leq L_{s}^{\max},
\quad \forall s \in \{1,\ldots,S\}.
\label{eq:opt_c}
\end{align}
\end{subequations}
where, $\widehat{R}_{s}^{-}(v,\mathbf{x}_t)$ and
$\widehat{L}_{s}^{+}(v,\mathbf{x}_t)$ denote conservative lower and upper
estimates of the predicted throughput and latency, respectively, for slice
$s$ under candidate value $v$. The weights $w_R,w_L \in [0,1]$ are fixed
design weights satisfying $w_R+w_L=1$. The normalized throughput and latency
gains are defined as, $\Delta_R(v;\mathbf{x}_t)
=
\frac{\widehat{R}(v,\mathbf{x}_t)
-\widehat{R}(v_0,\mathbf{x}_t)}
{\max(\widehat{R}(v_0,\mathbf{x}_t),\,\epsilon)}$,
and
$
\Delta_L(v;\mathbf{x}_t)
=
\frac{\widehat{L}(v_0,\mathbf{x}_t)
-\widehat{L}(v,\mathbf{x}_t)}
{\max(\widehat{L}(v_0,\mathbf{x}_t),\,\epsilon)}$. $\epsilon > 0$ is a small constant that prevents division by zero when the
baseline KPI estimate is negligibly small.

The objective function in~\eqref{eq:opt_a} captures the normalized QoS gain
obtained by replacing the current value $v_0$ with candidate value $v$.
The first term rewards throughput improvement, while the second term rewards
latency reduction and penalizes latency increases. The weights $w_R$ and
$w_L$ allow the operator to specify the relative importance of throughput
and latency. For example, larger $w_R$ values favor eMBB-oriented operation,
while larger $w_L$ values favor latency-sensitive URLLC operation.
Constraints~\eqref{eq:opt_b} and~\eqref{eq:opt_c} enforce slice-level SLA
feasibility under the candidate value, ensuring that the selected parameter
value does not reduce any active slice below its minimum throughput
requirement or increase its latency above the maximum allowable threshold.

The problem in~\eqref{eq:opt} presents two coupled challenges. First, before sufficient network observations have been collected, the
historical KPI database may contain too few entries to reliably estimate
$\widehat{R}_{s}(v,\mathbf{x}_t)$, $\widehat{L}_{s}(v,\mathbf{x}_t)$, or
their conservative bounds for candidate values $v \in \mathcal{V}_m$. Second, even after the database is populated, network
conditions may drift over time due to changing traffic patterns, user
mobility, and channel variations, making stored KPI estimates outdated. To
address these challenges, we propose \emph{SlicePlanner}, a two-phase
framework detailed in Sect.~\ref{sec:proposed_work}. The offline phase
pre-populates the historical KPI database under controlled configuration
trials, while the online control loop continuously updates the database,
validates configuration decisions, and re-triggers offline exploration when
persistent lack of QoS improvement is detected.
\begin{figure*}[ht!]
    \centering
\includegraphics[width=0.96\textwidth]{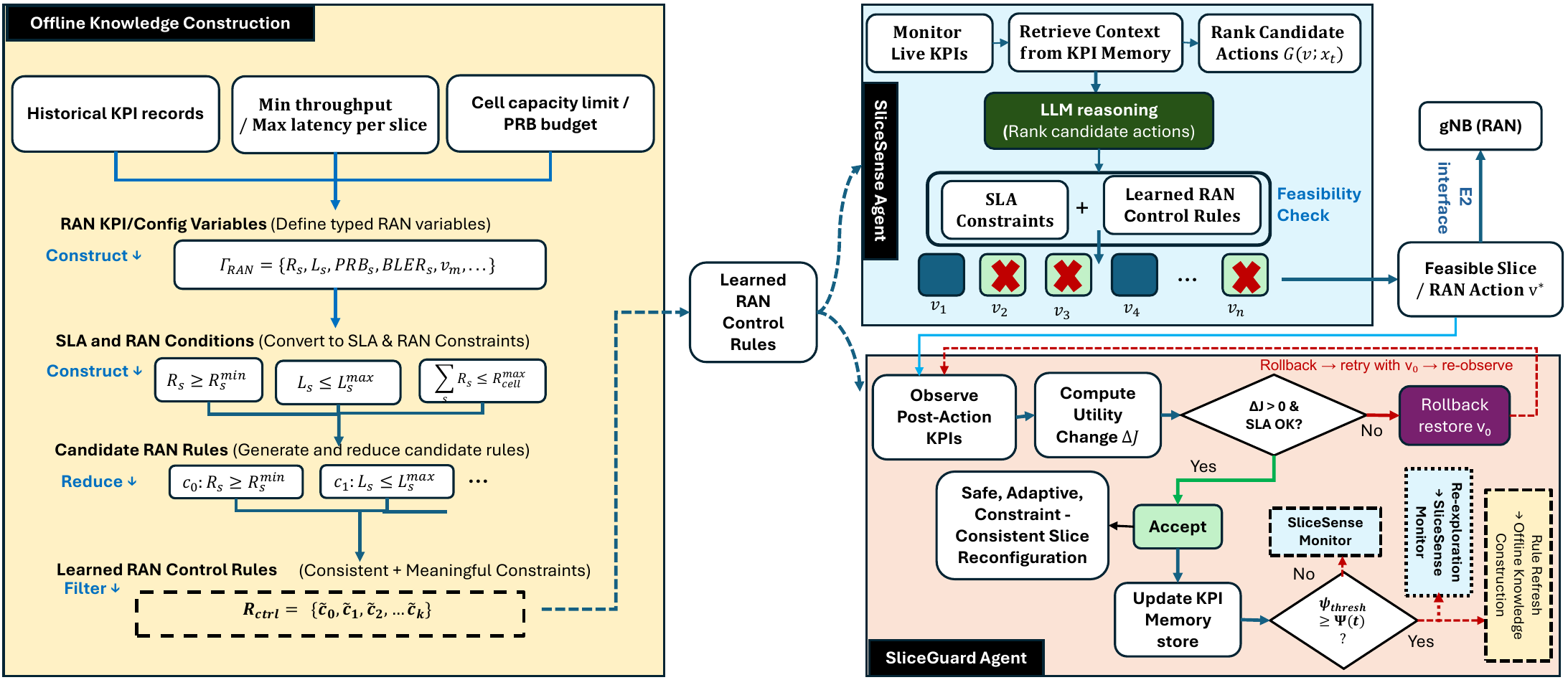}
    \caption{SlicePlanner logic-guided control pipeline. Offline knowledge construction uses historical KPI records and SLA rules to derive learned RAN control rules. During online operation, SliceSense monitors live
    KPIs, retrieves context from KPI memory, ranks candidate slice/RAN
    actions, and admits only actions that satisfy SLA constraints and learned
    control rules. SliceGuard validates the post-action KPI response, accepts
    effective updates, rolls back harmful changes, and triggers memory updates
    or re-exploration when rule inconsistency indicates drift.}
    \label{fig:PS}
    \vspace{-0.5cm}
\end{figure*}
\section{Proposed Solution: SlicePlanner}
\label{sec:proposed_work}
As shown in Fig.~\ref{fig:PS}, the framework consists of three components: offline knowledge construction, online decision generation
through the SliceSense Agent, and post-action validation through the SliceGuard Agent. Candidate slice/RAN actions are first generated from live KPI observations and historical context, then checked
against SLA constraints and learned RAN control rules before execution, and retained only if post-action KPIs confirm their benefit. This design is motivated by recent neurosymbolic networking systems, which show that ML-based networking pipelines can become more reliable and controllable when domain rules are represented as explicit constraints rather than being learned implicitly from data alone~\cite{he2025just, gong2024zoom2net}. In SlicePlanner, the learned rule set is denoted by $\mathcal{R}_{\mathrm{ctrl}}$ and represents data-consistent and semantically meaningful RAN control rules derived from historical KPI
records and operator-specified SLA rules.

At each control interval $t$, SlicePlanner receives the observed network context $\mathbf{x}_t$, which includes slice-level KPIs, active UE information, traffic demand, current configuration state, and recent control history. Given a RAN parameter $p_m$ with feasible candidate values $v \in \mathcal{V}_m$, SlicePlanner evaluates candidate actions using the predicted KPI response under the current context. Unlike data-driven controllers, SlicePlanner does not directly execute the
highest-gain candidate. Instead, each candidate is selected only if it is predicted to improve performance, satisfies operator-specified SLA
constraints, and does not conflict with the learned RAN control rules
$\mathcal{R}_{\mathrm{ctrl}}$.

\underline{\emph{\textbf{Offline Knowledge Construction.}}}
Fig.~\ref{fig:PS} includes the offline knowledge construction process. The goal of this stage is to convert raw operational
records into a compact set of learned RAN control rules that can be used during online slice reconfiguration. The input consists of historical KPI
records and operator-defined SLA rules. Each historical record has the form, $d_i = (\mathbf{x}_i, v_i, \mathbf{y}_i)$, where $\mathbf{x}_i$ denotes the observed network context, $v_i$ is the applied slice or RAN configuration value, and $\mathbf{y}_i$ contains the resulting KPI outcomes, including throughput, latency, PRB utilization, BLER, packet loss, jitter, and bitrate. (\textcircled{\small 1}) SlicePlanner defines a set of typed RAN KPI and configuration variables, denoted by $\Gamma_{\mathrm{RAN}} =
    \{R_s,L_s,\mathrm{PRB}_s,\mathrm{BLER}_s,v_m,\ldots\}$, where $R_s$, $L_s$, $\mathrm{PRB}_s$, and $\mathrm{BLER}_s$ represent measured slice-level KPIs, and $v_m$ denotes a controllable RAN parameter such as bandwidth, PRB allocation, or scheduling weight.
(\textcircled{\small 2}) The typed variables are converted into SLA and RAN conditions, including operator-specified SLA conditions defined as,

    $R_s \geq R_s^{\min},
    \qquad
    L_s \leq L_s^{\max}$, as well as resource and safety conditions, defined as,
$\sum_{s=1}^{S} R_s \leq R_{\mathrm{cell}}^{\max}$.
where $R_{\mathrm{cell}}^{\max}$ denotes the maximum aggregate cell throughput capacity.
These conditions specify whether a candidate control decision satisfies service requirements and shared RAN resource limits. (\textcircled{\small 3}) SlicePlanner constructs candidate RAN rules from these conditions, including throughput and latency feasibility, aggregate cell-capacity limits, bandwidth-throughput coupling, and cross-slice contention constraints. The candidate rules are then reduced and filtered to retain only those consistent with observed KPI records and semantically meaningful for RAN control. The learned RAN control rule set is denoted by,
$\mathcal{R}_{\mathrm{ctrl}}
    =
    \{\tilde{c}_0,\tilde{c}_1,\ldots,\tilde{c}_k\}$, where each $\tilde{c}_i$ is a retained control rule. Equivalently,
$\mathcal{R}_{\mathrm{ctrl}}$ can be interpreted as a conjunction of
learned constraints, $\mathcal{R}_{\mathrm{ctrl}}
    =
    \bigwedge_{\tilde{c}_i \in \mathcal{C}^{\prime}} \tilde{c}_i
    \label{eq:control_rule_conjunction}$, where $\mathcal{C}^{\prime}$ denotes the subset of strongest, consistent,
and meaningful constraints retained after rule construction and filtering. This offline stage addresses the limitation of using only manually written
rules. Operator-defined SLA rules capture essential service requirements,
but they may not cover deployment-specific relationships among load,
resource usage, channel quality, configuration history, and post-action KPI
responses. By mining learned RAN control rules from historical records,
SlicePlanner expands the rule coverage used during online control while
retaining interpretability.

\underline{\emph{\textbf{KPI-Response Prediction.}}} SliceSense uses a candidate-conditioned KPI-response surrogate to estimate
the effect of a RAN configuration before it is applied to the live network.
Let \(\mathbf{X}_{t-K+1:t}\) denote the recent observation window over the
last \(K\) control intervals, including slice KPIs, UE activity, traffic
demand, and configuration history. Given this context, the current
configuration \(\mathbf{v}(t)\), and a candidate value
\(v\in\mathcal{V}_m\), the surrogate predicts the post-action KPI response,
\begin{equation}
\{\widehat{\mathbf{y}}_{s,t}(v)\}_{s=1}^{S}
=
\mathcal{M}_{\theta}
\left(
\mathbf{X}_{t-K+1:t},\mathbf{v}(t),v
\right),
\label{eq:kpi_surrogate}
\end{equation}
where $\widehat{\mathbf{y}}_{s,t}(v)
=
[
\widehat{R}_{s}^{-},
\widehat{R}_{s},
\widehat{R}_{s}^{+},
\widehat{L}_{s}^{-},
\widehat{L}_{s},
\widehat{L}_{s}^{+}
]$ contains the predicted lower, nominal, and upper estimates of throughput and latency for slice $s$ under candidate value $v$.
The mean estimates \(\widehat{R}_{s}\) and \(\widehat{L}_{s}\) are used for
candidate ranking, while the conservative bounds
\(\widehat{R}_{s}^{-}\) and \(\widehat{L}_{s}^{+}\) are used for
SLA-admissibility checking.
The surrogate represents the RAN state as a temporal bipartite graph. At
control interval \(k\), SlicePlanner constructs
\(\mathcal{G}_k=(\mathcal{U}_k\cup\mathcal{S},\mathcal{E}_k)\), where
\(\mathcal{U}_k\) denotes the set of active UE nodes, \(\mathcal{S}\) denotes
the set of slice nodes, and an edge \((u,s)\in\mathcal{E}_k\) indicates that
UE \(u\) is served by slice \(s\). Node, edge, and graph-level features include throughput, latency, packet loss,
BLER, jitter, PRB usage, traffic demand, configuration state, and channel
indicators. The candidate value \(v\) is encoded as an action
embedding and provided with the graph features. The model computes candidate-conditioned slice embeddings and then
aggregates them over time, 
\begin{figure*}[t]
    \centering
    \setlength{\tabcolsep}{2pt}
    \begin{tabular}{@{}cc@{}}
        \includegraphics[width=0.49\textwidth]{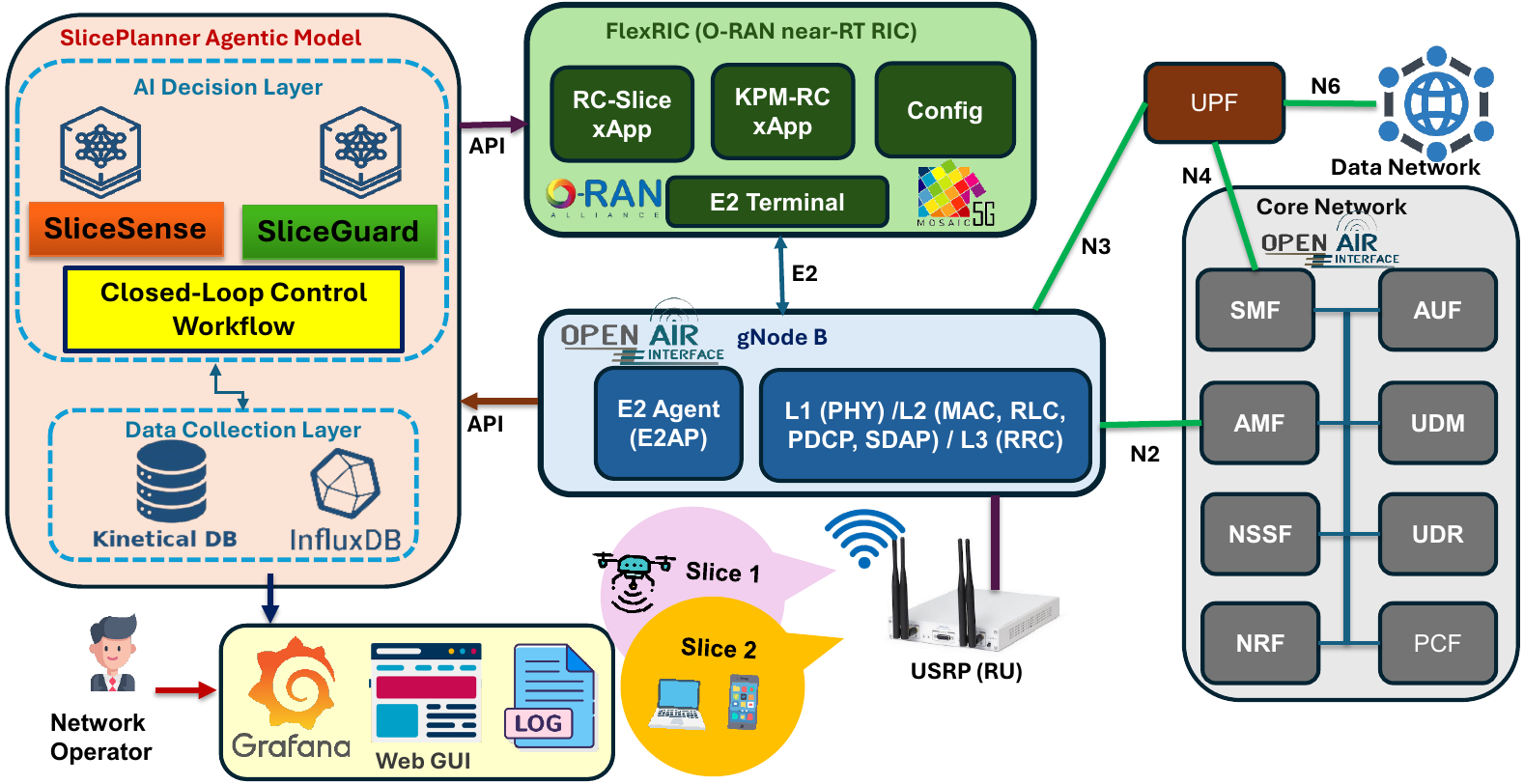} &
        \includegraphics[width=0.40\textwidth]{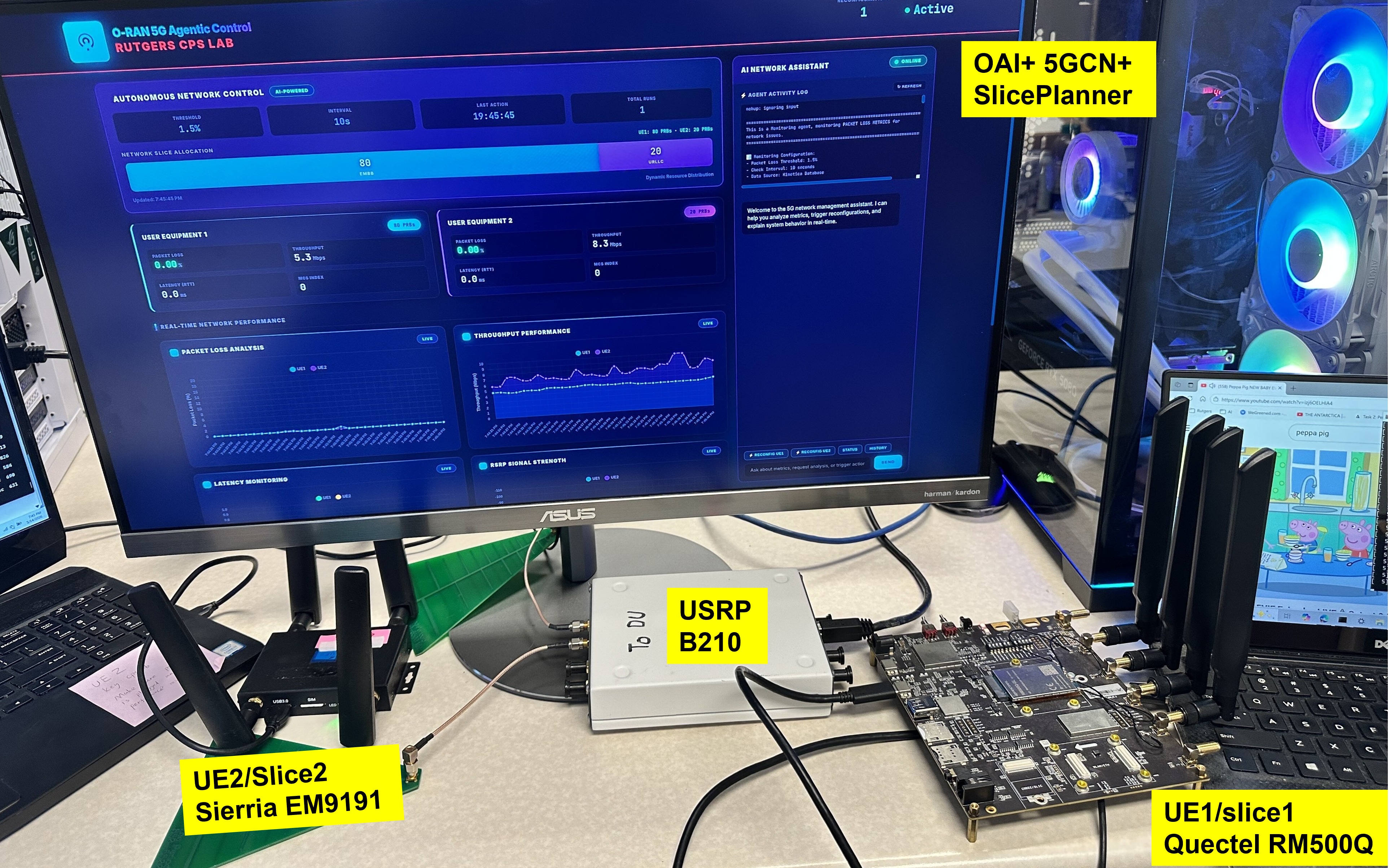} \\
        \small (a) & \small (b)
    \end{tabular}
    \caption{Over-the-air O-RAN testbed used to validate SlicePlanner: (a) logical architecture and (b) hardware testbed setup.}
    \label{fig:testbed}
    \vspace{-0.2in}
\end{figure*}
\begin{equation}
\begin{aligned}
\mathbf{h}_{s,k}(v)
&=
g_{\theta_g}
\left(
\mathcal{G}_k,\mathbf{x}_k,\mathbf{v}(k),v
\right), \\
\mathbf{z}_{s,t}(v)
&=
q_{\theta_q}
\left(
\mathbf{h}_{s,t-K+1}(v),\ldots,\mathbf{h}_{s,t}(v)
\right), \\
\widehat{\mathbf{y}}_{s,t}(v)
&=
o_{\theta_o}
\left(
\mathbf{z}_{s,t}(v)
\right).
\end{aligned}
\label{eq:surrogate_arch}
\end{equation}
Here, \(g_{\theta_g}\) is a graph encoder, \(q_{\theta_q}\) is a temporal
encoder, and \(o_{\theta_o}\) outputs the mean and quantile KPI estimates.

The surrogate is trained on historical transition records
\((\mathbf{X}_{i-K+1:i},\mathbf{v}(i),v_i,\mathbf{y}_{i+\delta})\), where
\(\mathbf{y}_{i+\delta}\) is the measured KPI response after a settling
window \(\delta\). The training objective combines mean prediction error
with quantile loss,
\begin{equation}
\mathcal{L}(\theta)
=
\sum_{i,s}
\left[
\left\|
\mathbf{y}_{s,i+\delta}
-
\widehat{\boldsymbol{\mu}}_{s,i}
\right\|_2^2
+
\lambda_q
\sum_{\tau\in\mathcal{Q}}
\rho_{\tau}
\left(
\mathbf{y}_{s,i+\delta}
-
\widehat{\mathbf{y}}_{s,i}^{(\tau)}
\right)
\right],
\label{eq:surrogate_loss}
\end{equation}
where
\(\mathbf{y}_{s,i+\delta}=[R_{s,i+\delta},L_{s,i+\delta}]\),
\(\widehat{\boldsymbol{\mu}}_{s,i}=[\widehat{R}_{s,i},\widehat{L}_{s,i}]\),
\(\rho_{\tau}(\cdot)\) is the componentwise pinball loss, and
\(\mathcal{Q}\) contains the lower and upper quantile levels used to form
uncertainty bounds. During online operation, \(\mathcal{M}_{\theta}\) is used only to score
candidate values in the discrete set \(\mathcal{V}_m\). 
\underline{\emph{\textbf{LLM-Assisted Reasoning in SliceSense.}}}
SliceSense uses an instruction-tuned Llama-3.1 model served through NVIDIA NIM
as an orchestration layer within the closed-loop control pipeline. At each
control interval, the LLM receives a structured prompt containing the current
network context $\mathbf{x}_t$, candidate actions ranked by
$\mathcal{M}_{\theta}$, and the learned RAN control rules
$\mathcal{R}_{\mathrm{ctrl}}$ formatted as natural-language constraints.
Given this input, the LLM retrieves relevant KPI context from Kinetica,
reasons over the candidate ranking to identify the most appropriate
reconfiguration, and formats the selected candidate as a structured control
decision. The LLM does not predict numerical KPIs or issue unchecked RAN
commands. A proposed candidate is retained only after passing
surrogate-based scoring, SLA admissibility checking, rule-consistency
verification via~\eqref{eq:smt_check}, and SliceGuard post-action validation.
\indent\underline{\emph{\textbf{SliceSense: Logic-Guided Selection.}}}
 As shown in Fig.~\ref{fig:PS}, SliceSense performs three operations at each control interval: monitoring live KPIs, retrieving
context from KPI memory, and ranking candidate actions. SliceSense first monitors live KPIs and detects QoS degradation. For each slice $s$, it checks whether the observed throughput and latency
satisfy the SLA conditions, defined as, $\bar{R}_s(t) \geq R_s^{\min},
    \qquad
    \bar{L}_s(t) \leq L_s^{\max}$. When a violation or persistent degradation is detected, SliceSense
retrieves the relevant recent context from the KPI memory, including the
affected slice or UE, recent KPI trends, the current configuration
$\mathbf{v}(t)$, previous control actions, and historical outcomes under
similar contexts.
\begin{figure*}[t]
    \centering
    \setlength{\tabcolsep}{2pt}
    \begin{tabular}{@{}ccc@{}}
        \includegraphics[width=0.31\textwidth,height=1.1in]{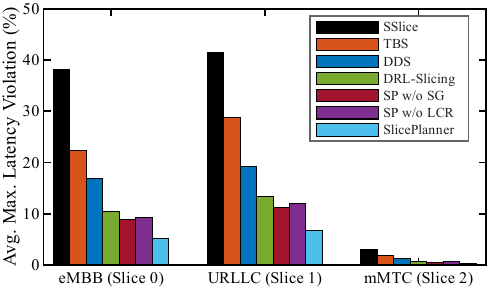} &
        \includegraphics[width=0.31\textwidth,height=1.1in]{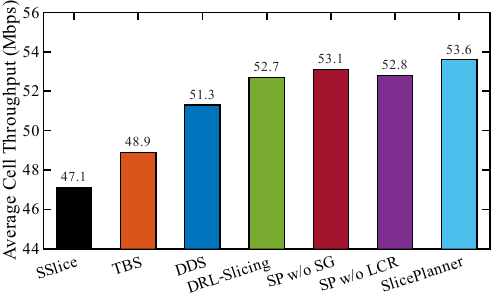} &
        \includegraphics[width=0.31\textwidth,height=1.1in]{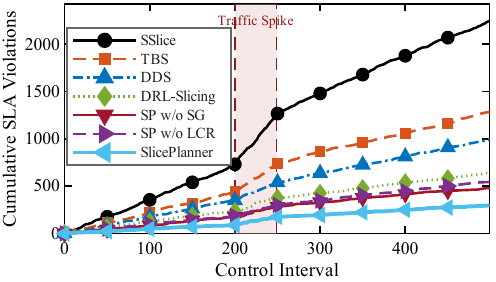} \\
        \small (a) & \small (b) & \small (c)
    \end{tabular}
    \vspace{-0.06in}
    \caption{Numerical simulation results comparing SlicePlanner against baseline methods: (a) slice-level average maximum latency SLA violation rate across eMBB, URLLC, and mMTC slices; (b) average cell throughput; and (c) cumulative SLA violations over 500 control intervals, with a traffic spike injected.}
    \label{fig:latency}
    \vspace{-0.05in}
\end{figure*}
\begin{figure*}[!t]
    \centering
    \setlength{\tabcolsep}{0pt}
    \begin{tabular}{@{}cccc@{}}
        \includegraphics[width=0.232\textwidth]{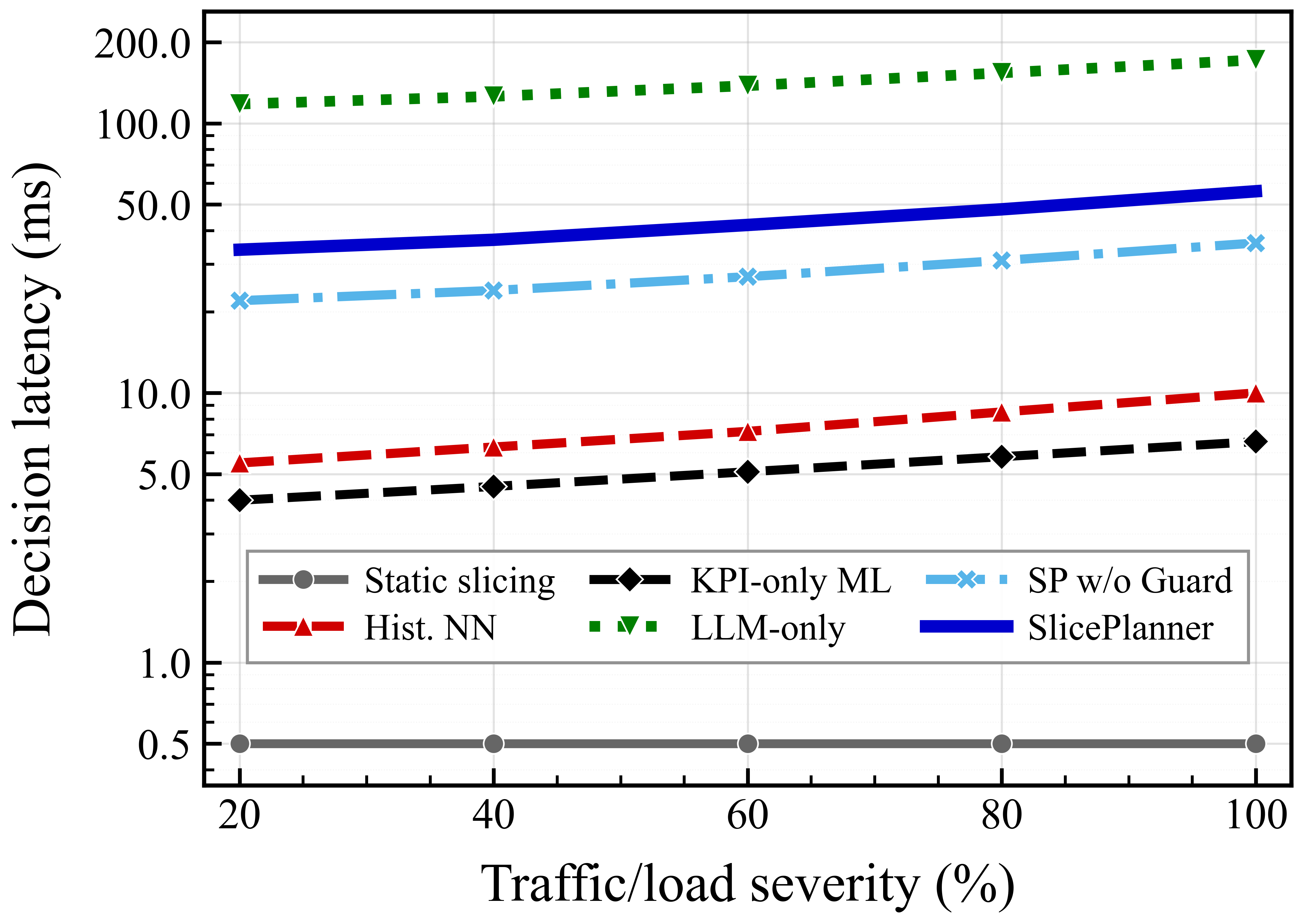} &
        \includegraphics[width=0.232\textwidth]{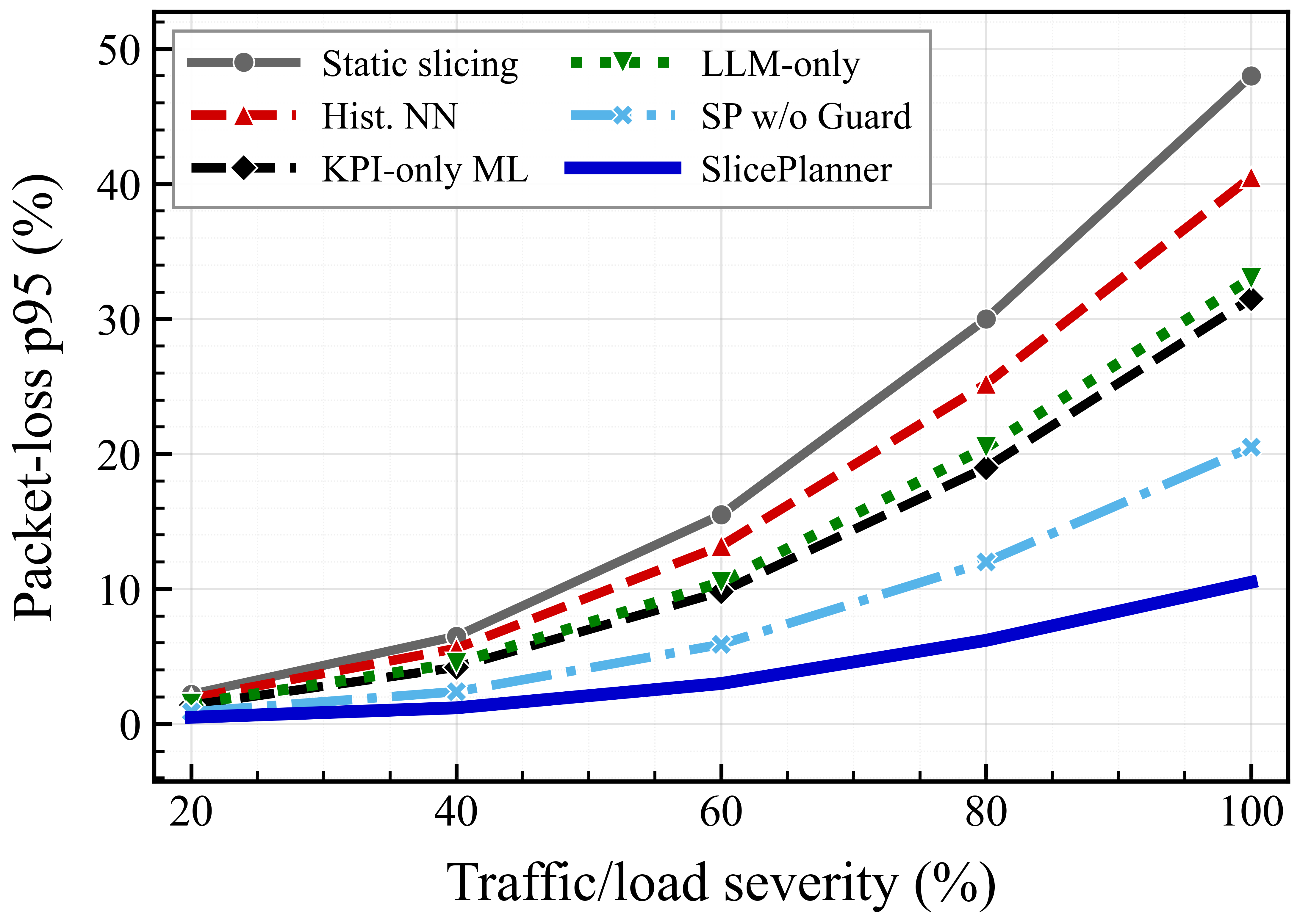} &
        \includegraphics[width=0.232\textwidth]{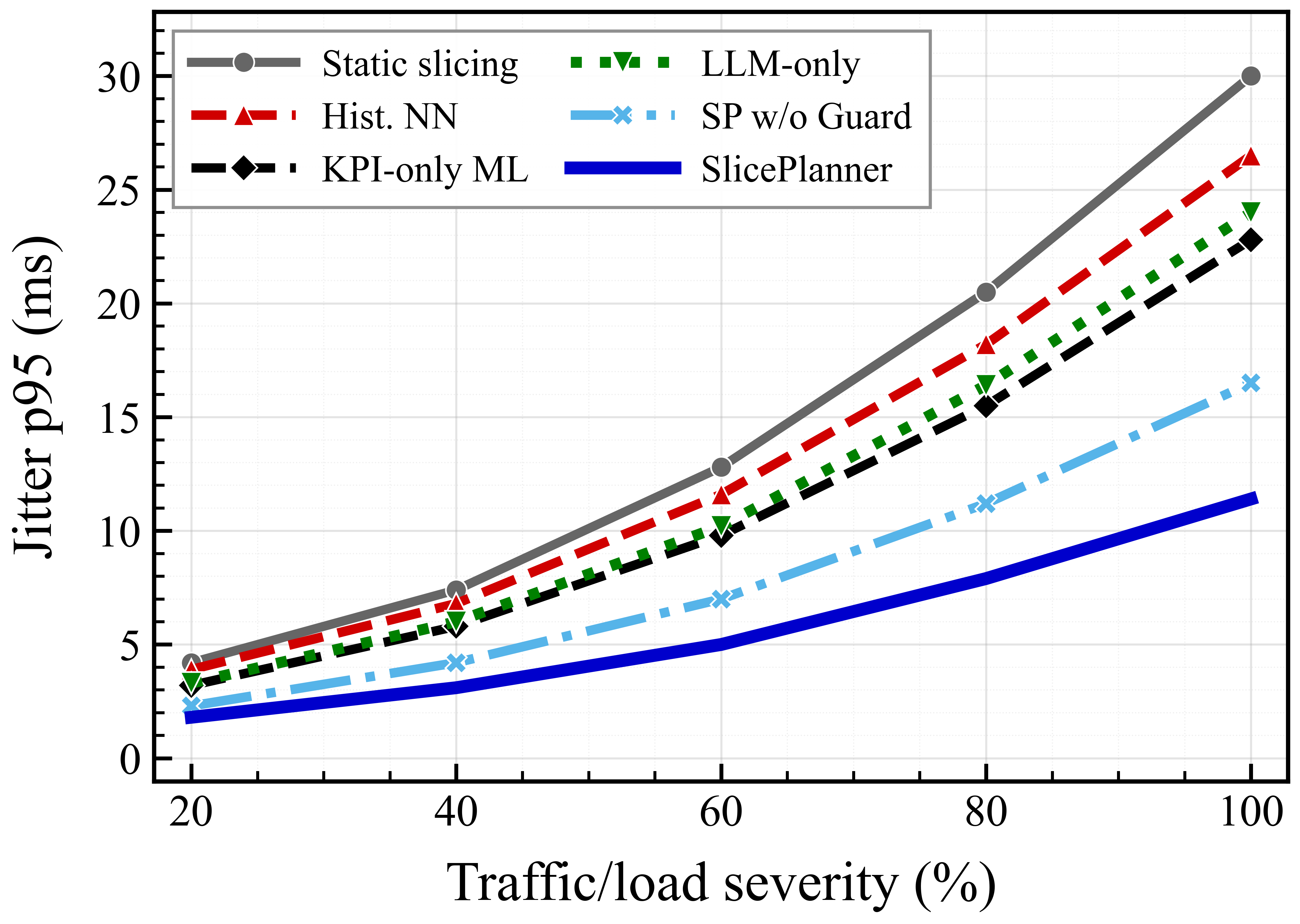} &
        \includegraphics[width=0.232\textwidth]{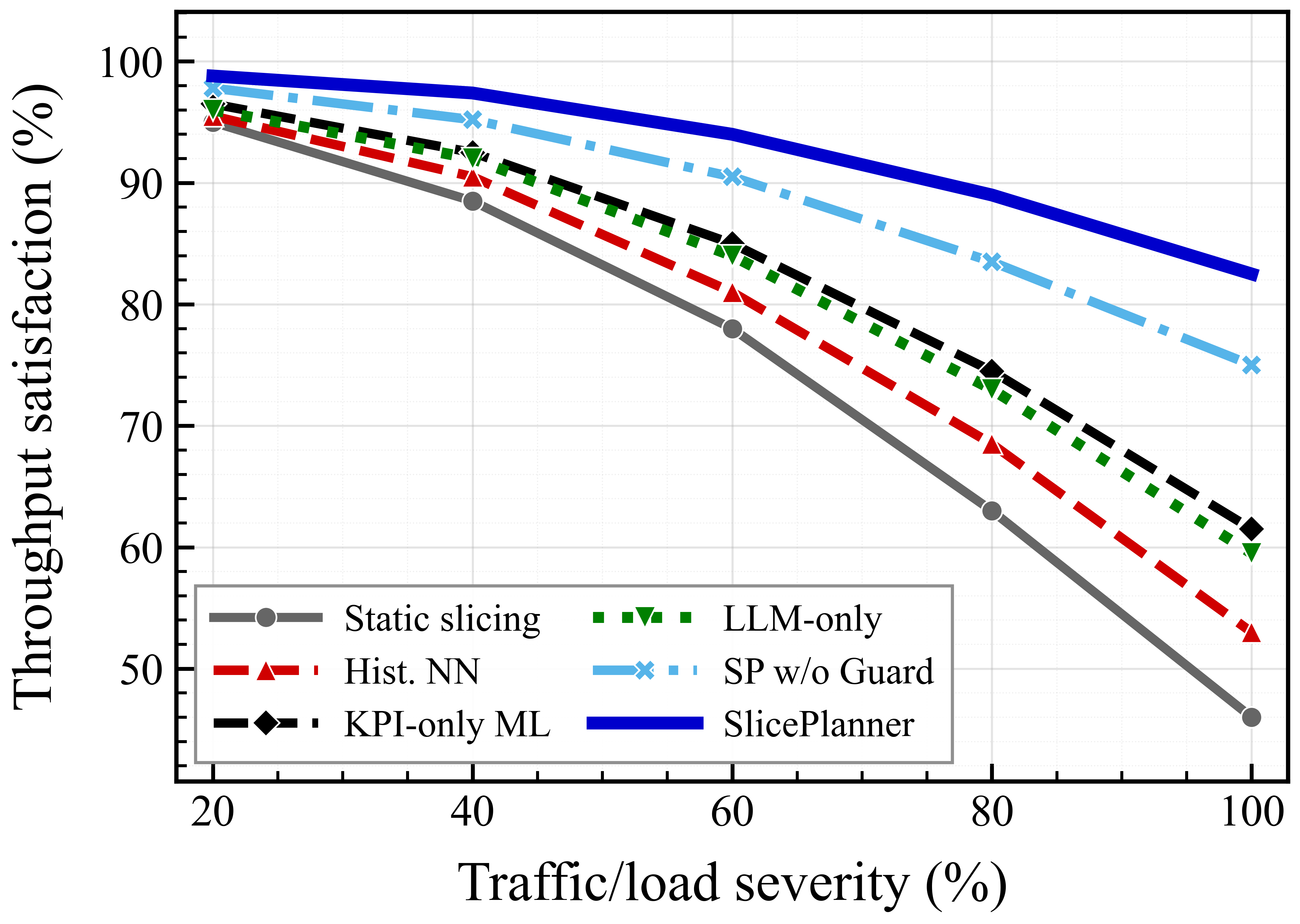} \\
        \small (a) & \small (b) & \small (c) & \small (d)
    \end{tabular}
    \caption{Over-the-air O-RAN testbed performance under increasing traffic/load severity: (a) decision latency, (b) packet-loss p95, (c) jitter p95, and (d) throughput satisfaction. Lower values are better for decision latency, packet loss, and jitter, while higher throughput satisfaction indicates better service quality. SlicePlanner provides the best overall tradeoff, maintaining low tail degradation and high throughput satisfaction with moderate decision overhead.}
    \label{fig:performance_load_severity_urll}
    \vspace{-0.18in}
\end{figure*}
For each candidate value \(v \in \mathcal{V}_m\), SliceSense queries the KPI-response surrogate \(\mathcal{M}_{\theta}\), which predicts post-action throughput and latency for each slice with uncertainty bounds. SliceSense uses the mean predictions to compute the normalized gain \(G(v;\mathbf{x}_t)\), and uses conservative bounds to determine whether the candidate is safe for execution. Candidate actions are then ranked by \(G(v;\mathbf{x}_t)\), allowing historical network responses to guide the action most likely to improve the current network state.

Before execution, SliceSense applies an SLA-aware feasibility check. This check
has two parts. The first part enforces explicit SLA feasibility, defined as,
\begin{equation}
    \widehat{R}_{s}^{-}(v,\mathbf{x}_t) \geq R_s^{\min},
    \qquad
    \widehat{L}_{s}^{+}(v,\mathbf{x}_t) \leq L_s^{\max}.
    \label{eq:admissibility_sla}
\end{equation}
The second part checks consistency with the learned RAN control rules
$\mathcal{R}_{\mathrm{ctrl}}$. For a candidate value $v$, current
context $\mathbf{x}_t$, and predicted KPI vector
$\hat{\mathbf{y}}=[\widehat{R}_s,\widehat{L}_s]_{s=1}^{S}$, SlicePlanner
encodes the candidate decision as logical facts
$\Phi(v,\mathbf{x}_t,\hat{\mathbf{y}})$ and evaluates as,
\begin{equation}
    f_{\mathrm{check}}(v,\mathbf{x}_t,\hat{\mathbf{y}})
    =
    \operatorname{SAT}\!\left(
        \mathcal{R}_{\mathrm{ctrl}}
        \wedge
        \Phi(v,\mathbf{x}_t,\hat{\mathbf{y}})
    \right).
    \label{eq:smt_check}
\end{equation}
where $\operatorname{SAT}(\cdot)$ denotes a satisfiability check. If the formula in~\eqref{eq:smt_check} is satisfiable, the candidate is
consistent with the learned RAN control rules. If it is unsatisfiable, the
candidate violates at least one learned rule and is rejected. In our implementation, predicted KPIs are represented as
uncertainty-aware intervals,
\begin{equation}
\begin{aligned}
\widehat{R}_s(v,\mathbf{x}_t)
&\in
\left[
\widehat{R}_s^{-}(v,\mathbf{x}_t),
\widehat{R}_s^{+}(v,\mathbf{x}_t)
\right], \\
\widehat{L}_s(v,\mathbf{x}_t)
&\in
\left[
\widehat{L}_s^{-}(v,\mathbf{x}_t),
\widehat{L}_s^{+}(v,\mathbf{x}_t)
\right].
\end{aligned}
\label{eq:kpi_intervals}
\end{equation}
This allows SlicePlanner to account for estimation uncertainty before execution. The feasible action set at time $t$ is therefore defined as,
\begin{equation}
\begin{aligned}
\mathcal{A}_t =
\big\{ v \in \mathcal{V}_m \mid
& \widehat{R}_{s}^{-}(v,\mathbf{x}_t) \geq R_s^{\min},
\ \widehat{L}_{s}^{+}(v,\mathbf{x}_t) \leq L_s^{\max}, \\
& \operatorname{SAT}\!\big(
\mathcal{R}_{\mathrm{ctrl}}\wedge
\Phi(v,\mathbf{x}_t,\hat{\mathbf{y}})
\big)
\big\}.
\end{aligned}
\label{eq:admissible_set}
\end{equation}
SliceSense selects, $v^*
    =
    \arg\max_{v \in \mathcal{A}_t}
    G(v;\mathbf{x}_t)$.
If $\mathcal{A}_t$ is empty, SlicePlanner preserves the current
configuration $v_0$, records the case as unresolved degradation, and
triggers controlled re-exploration if the condition persists. Otherwise,
the selected feasible action $v^*$ is applied to the RAN through the
available slice or RAN reconfiguration interface.

\indent\underline{\emph{\textbf{SliceGuard: Post-Action Validation.}}} After the feasible action $v^*$ is applied,
SliceGuard waits for a configurable settling interval and then observes
the post-action KPIs. It compares the post-action response with the
pre-action baseline associated with the previous configuration $v_0$. Let $J(v)$ denote the utility of applying candidate value $v$ to the selected RAN parameter. For a throughput-latency objective, we define the utility as,
\begin{equation}
    J(v)
    =
    w_R
    \frac{\bar{R}(v)}
    {\max(\bar{R}_{\mathrm{ref}},\epsilon)}
    +
    w_L
    \left(
    1 -
    \frac{\bar{L}(v)}
    {\max(\bar{L}_{\mathrm{ref}},\epsilon)}
    \right),
    \label{eq:utility}
\end{equation}
where $\bar{R}(v)$ and $\bar{L}(v)$ are the measured aggregate throughput
and latency under configuration $v$, and
$\bar{R}_{\mathrm{ref}}$ and $\bar{L}_{\mathrm{ref}}$ are reference values
obtained from the pre-action state, historical baselines, or service-level
objectives. Given the previous configuration $v_0$ and the newly applied
configuration $v^*$, SliceGuard computes the post-action utility gain as, $\Delta J = J(v^*) - J(v_0)$. If $\Delta J>0$ and all active slices remain SLA-compliant, SliceGuard
accepts the update and commits the new record
$(\mathbf{x}_t,v^*,\mathbf{y}_t)$ to the KPI memory. If
$\Delta J \leq 0$, or if another slice experiences degradation, SliceGuard
rejects the update and rolls back to $v_0$. SliceGuard also checks whether the observed post-action KPI vector $\mathbf{y}_t$ remains consistent with the learned RAN control rules. It
tracks rule inconsistency through the sliding-window score, defined as,
\begin{equation}
\small
    \Psi(t)
    =
    \frac{1}{W}
    \sum_{k=t-W+1}^{t}
    \mathbf{1}\!\left[
        \operatorname{UNSAT}\!\left(
            \mathcal{R}_{\mathrm{ctrl}}
            \wedge
            \Phi(v_k,\mathbf{x}_k,\mathbf{y}_k)
        \right)
    \right],
    \label{eq:psi}
\end{equation}
where $\operatorname{UNSAT}(\cdot)$ denotes an unsatisfiability check. $W$ is the validation window length and $\mathbf{1}[\cdot]$ is the
indicator function. When $\Psi(t)\geq\psi_{\mathrm{thresh}}$, live
observations repeatedly contradict the learned RAN control rules, which
indicates that the current network distribution has drifted from the
conditions under which $\mathcal{R}_{\mathrm{ctrl}}$ was constructed.
SliceGuard then triggers memory update, controlled re-exploration, and
rule reconstruction.

\indent\underline{\emph{\textbf{Online Adaptation and Rule Refresh.}}} During online operation, the KPI memory is continuously updated with newly observed records. Accepted configurations enrich the memory with successful control examples, while rolled-back configurations provide negative evidence for future decisions. This feedback loop allows SlicePlanner to adapt to non-stationary traffic, changing channel conditions, UE mobility, and evolving service priorities. Rule refresh is triggered by repeated rollback events or a high rule-inconsistency score $\Psi(t)$. When triggered, SlicePlanner revisits the offline knowledge construction process using updated KPI memory. The learned RAN control rules $\mathcal{R}_{\mathrm{ctrl}}$ are refreshed to better reflect the current network distribution. This prevents SlicePlanner from relying indefinitely on outdated historical relationships while avoiding unnecessary re-exploration during stable operation.

\begin{figure*}[!t]
    \centering
    \setlength{\tabcolsep}{0pt}
    \begin{tabular}{@{}cccc@{}}
        \includegraphics[width=0.232\textwidth]{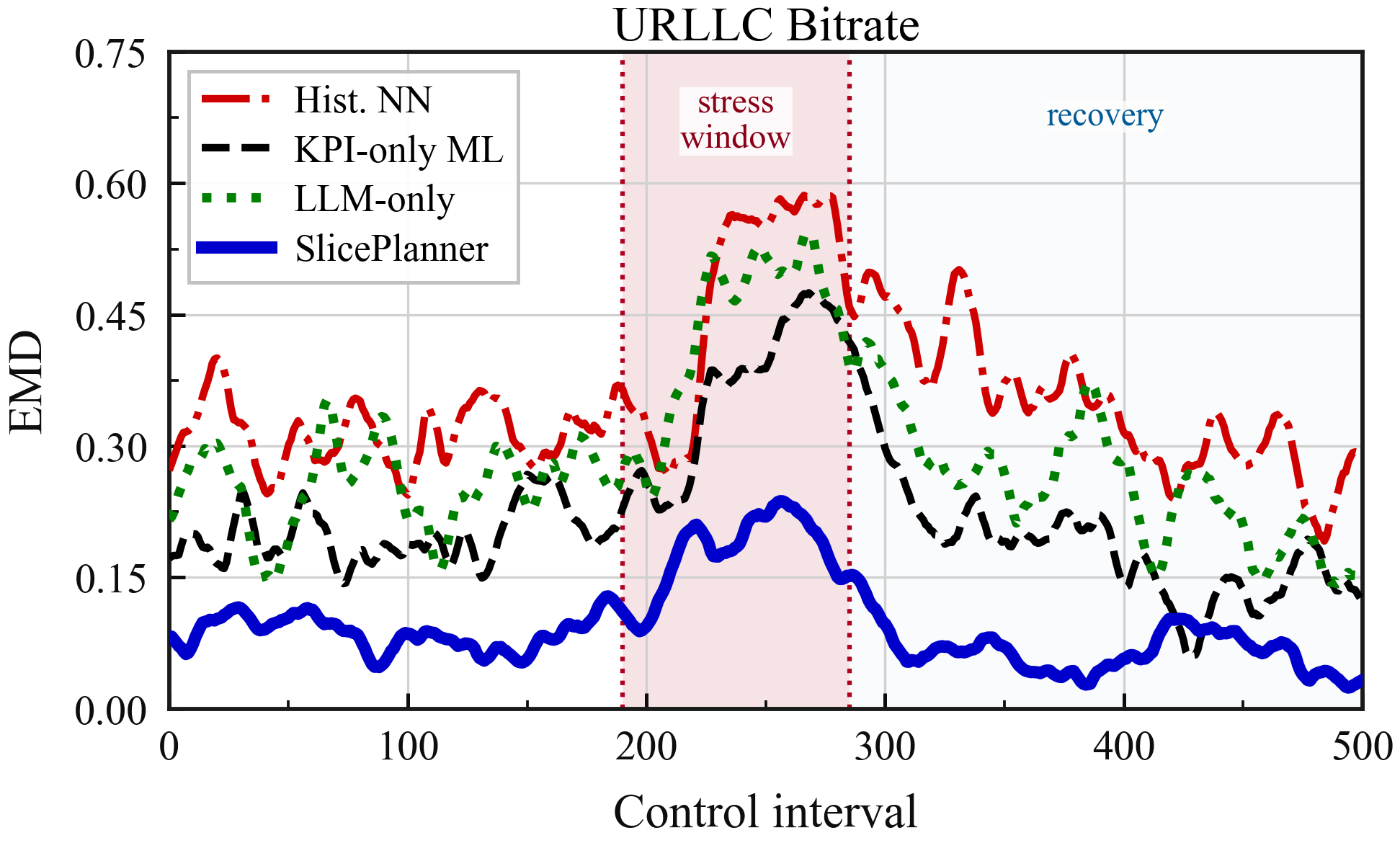} &
        \includegraphics[width=0.232\textwidth]{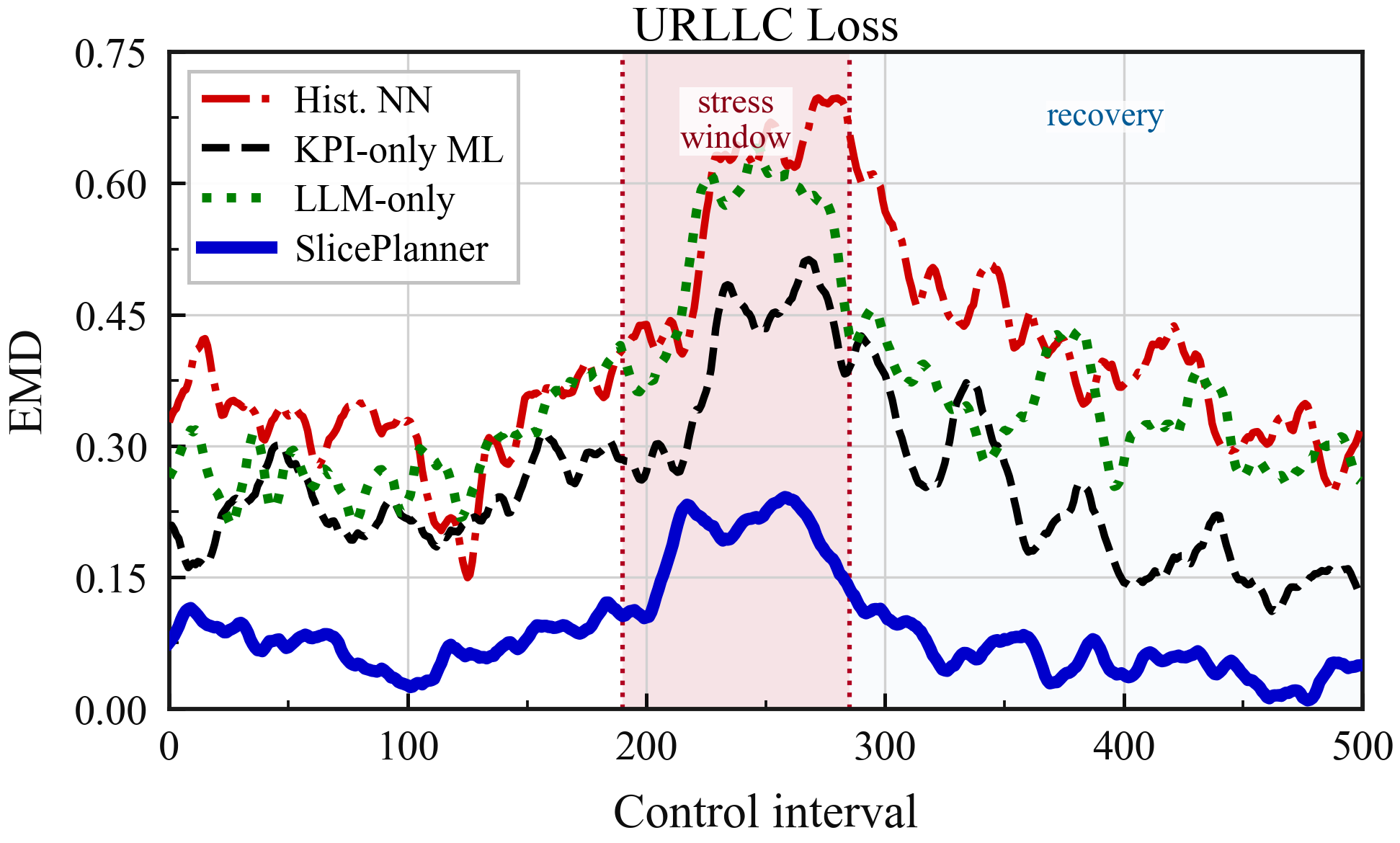} &
        \includegraphics[width=0.232\textwidth]{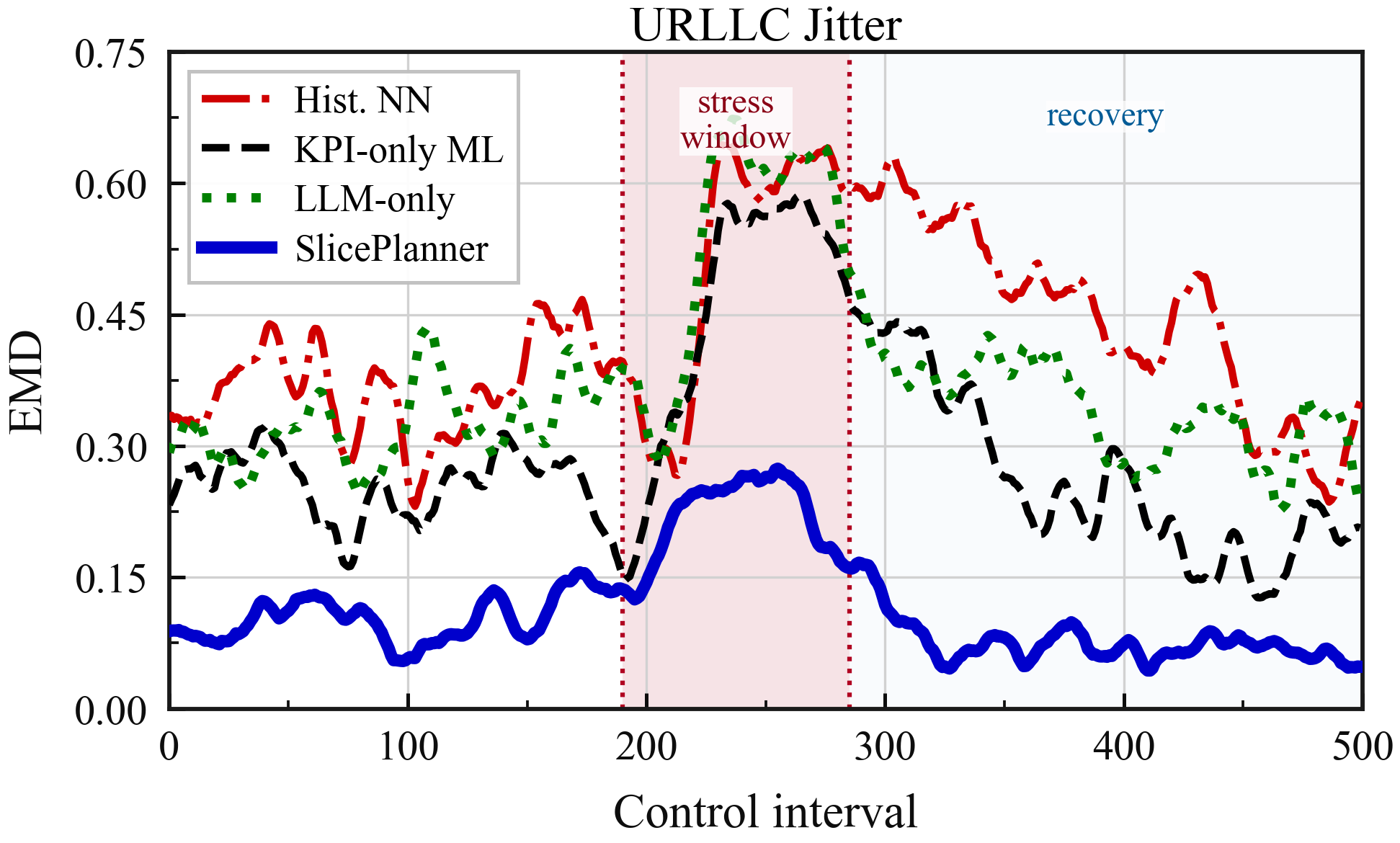} &
        \includegraphics[width=0.232\textwidth]{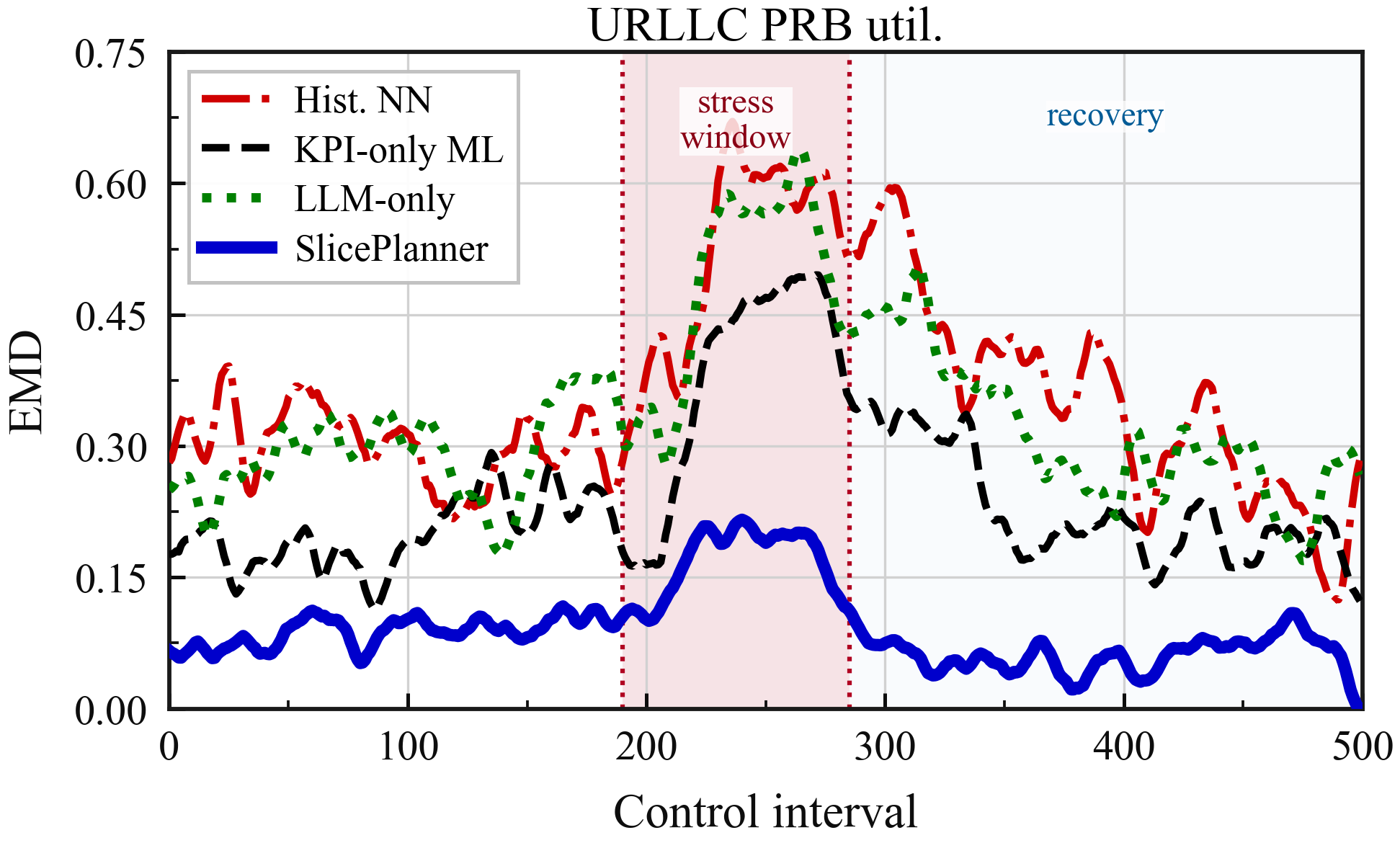} \\
        \small (a) URLLC bitrate &
        \small (b) URLLC packet loss &
        \small (c) URLLC jitter &
        \small (d) URLLC PRB utilization \\[0.04in]

        \includegraphics[width=0.232\textwidth]{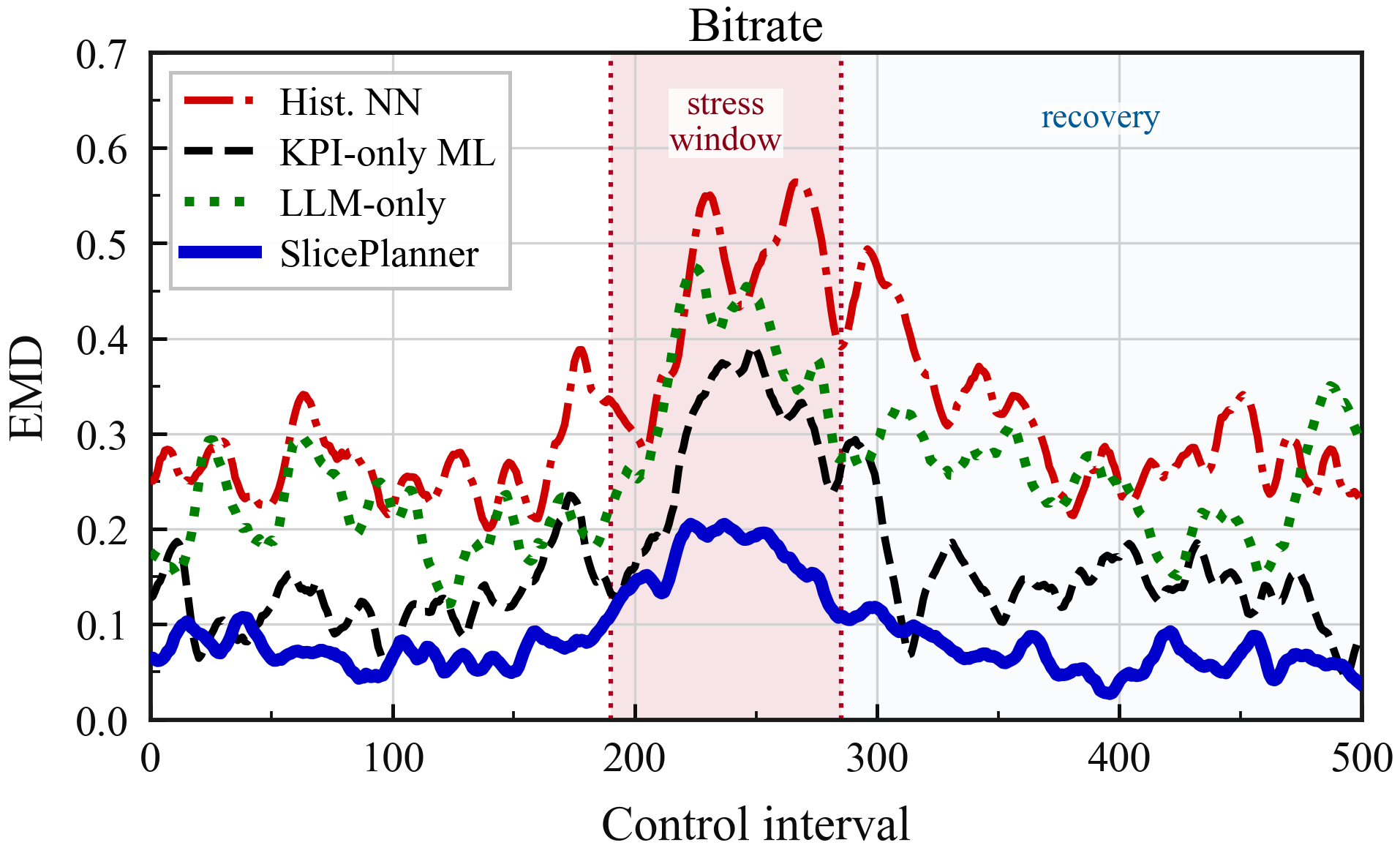} &
        \includegraphics[width=0.232\textwidth]{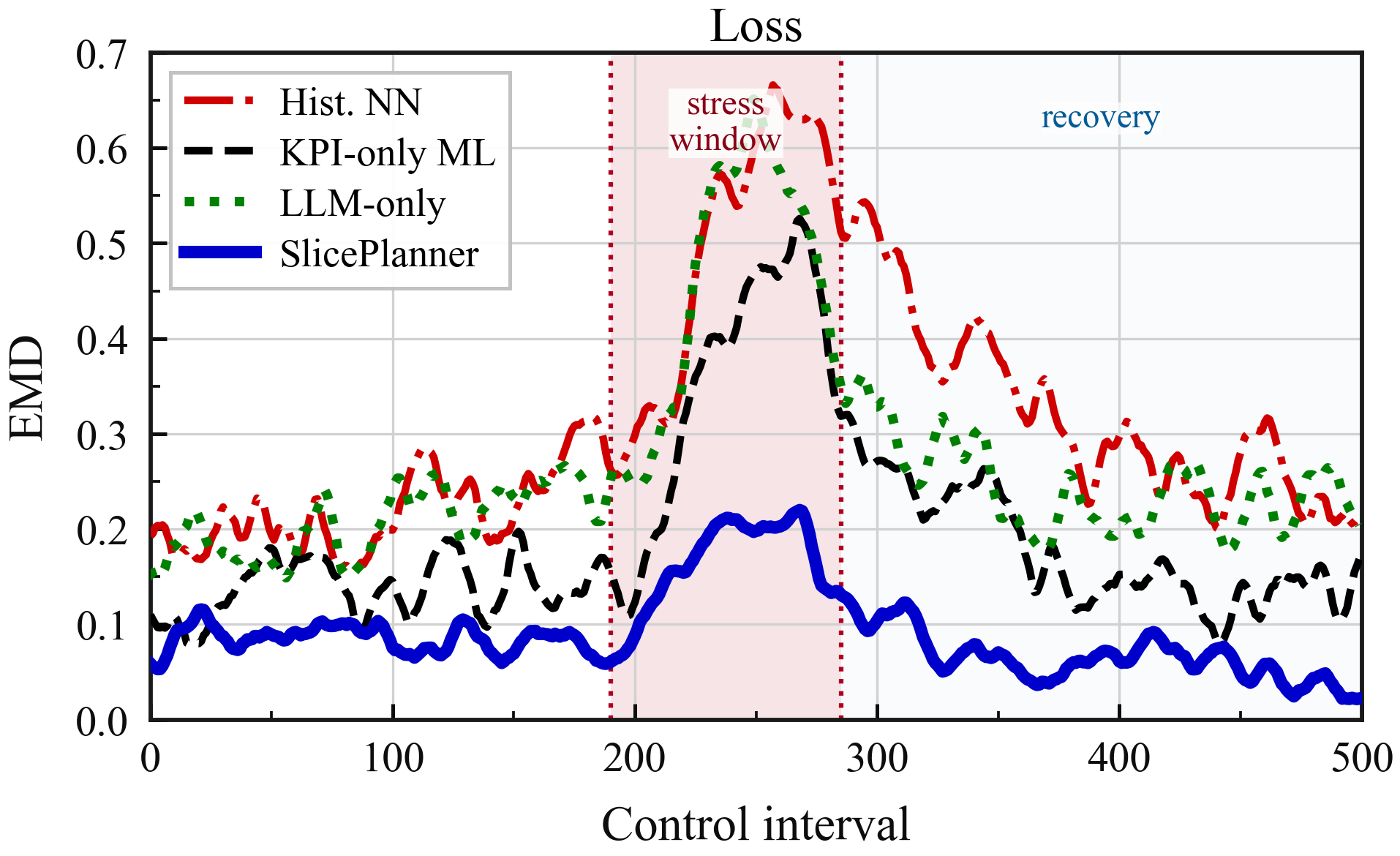} &
        \includegraphics[width=0.232\textwidth]{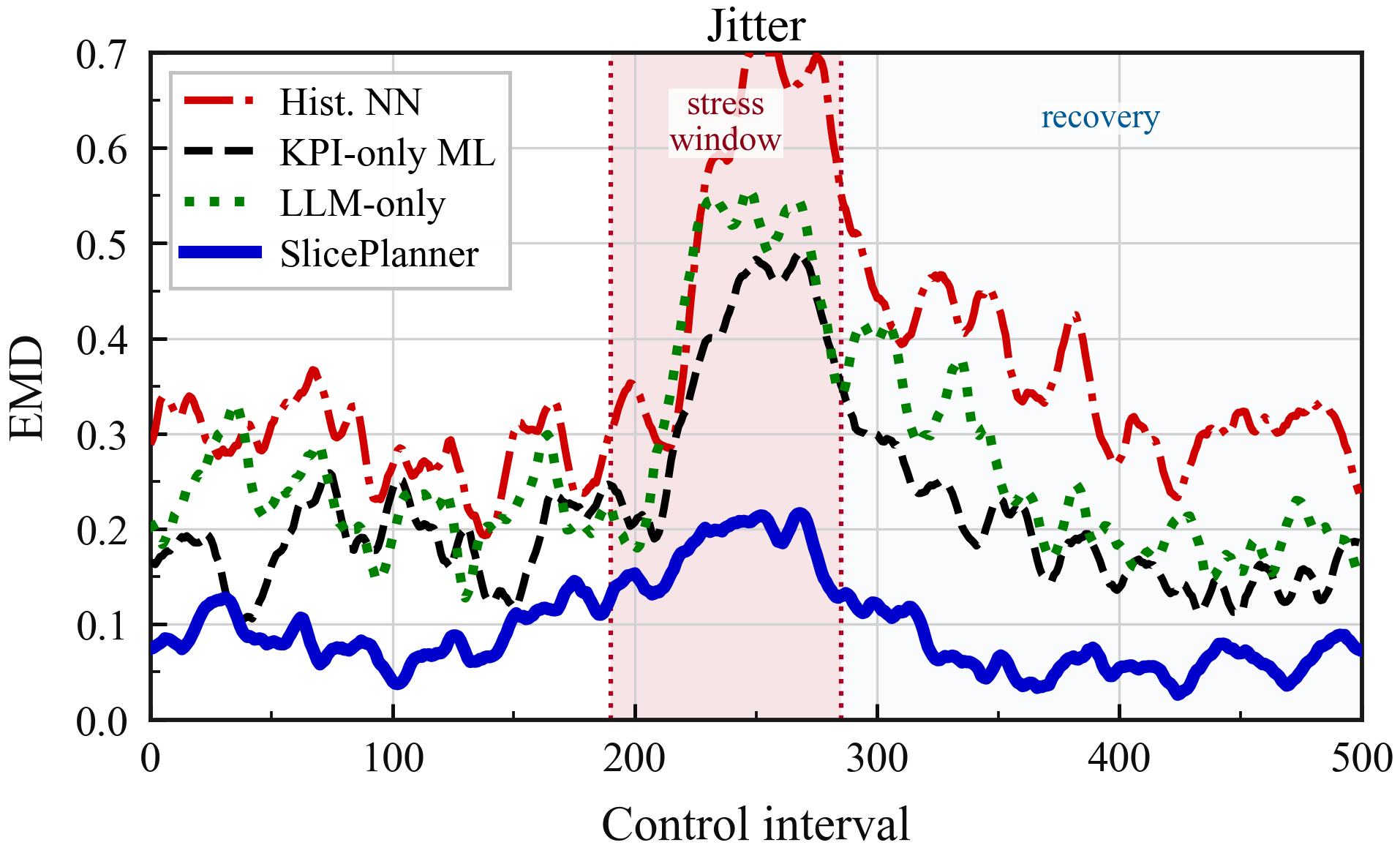} &
        \includegraphics[width=0.232\textwidth]{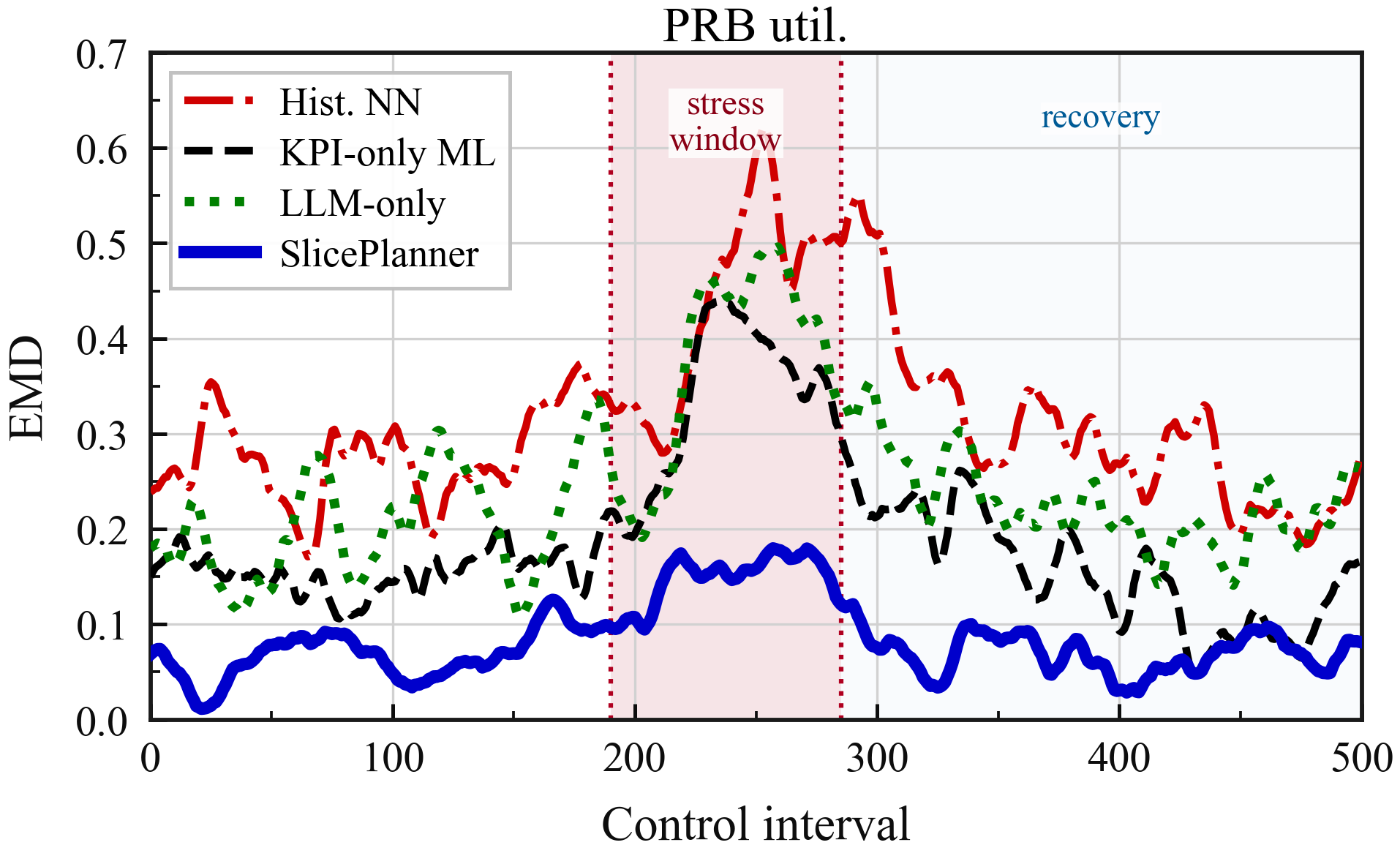} \\
        \small (e) eMBB bitrate &
        \small (f) eMBB packet loss &
        \small (g) eMBB jitter &
        \small (h) eMBB PRB utilization
    \end{tabular}
    \vspace{-0.06in}
    \caption{KPI-response fidelity during a temporary network-stress interval for the URLLC and eMBB slices. The first row reports the Earth Mover's Distance (EMD) for the URLLC slice across bitrate, packet loss, jitter, and PRB utilization, while the second row reports the corresponding EMD values for the eMBB slice. Lower EMD indicates closer agreement with the target KPI response. SlicePlanner achieves the lowest deviation across both slices and all KPI categories, indicating more stable and accurate closed-loop control under network stress.}
    \label{fig:emd_response_fidelity_urllc_embb}
    \vspace{-0.18in}
\end{figure*}
\section{Performance Evaluation}\label{sec:performance_eval}
We evaluate SlicePlanner through numerical simulations before deployment on the over-the-air O-RAN testbed. The purpose of the simulation study is to isolate the effect of logic-guided slice planning under controlled network conditions and compare SlicePlanner against representative static, rule-based, data-driven, and reinforcement-learning baselines.
\indent\textbf{Simulation Settings.}
We consider a single-cell O-RAN slicing scenario where the controller periodically allocates radio resources based on recent KPI measurements and slice-level SLA requirements. The cell serves $N=26$ UEs across $S=3$ slices: eMBB 
Slice~0 with 10 UEs and SLA target 
$(R_0^{\min},L_0^{\max})=(2~\mathrm{Mbps},100~\mathrm{ms})$; 
URLLC Slice~1 with 10 UEs and 
$(R_1^{\min},L_1^{\max})=(2~\mathrm{Mbps},10~\mathrm{ms})$; 
and mMTC Slice~2 with 6 UEs and 
$(R_2^{\min},L_2^{\max})=(2~\mathrm{Mbps},100~\mathrm{ms})$. 
The minimum throughput requirement is set uniformly to 
2~Mbps across all slices to isolate the effect of latency 
SLA differentiation in the evaluation, with URLLC imposing 
a tighter latency constraint of 10~ms compared to 
100~ms for eMBB and mMTC. Slice priorities are set to 
$\alpha_0=0.4$, $\alpha_1=0.4$, and $\alpha_2=0.2$, 
reflecting the higher service priority of eMBB and URLLC 
relative to mMTC, with $\sum_{s=0}^{2}\alpha_s=1$ 
satisfied by construction. The objective weights are fixed 
at $w_R=w_L=0.5$ to balance throughput and latency 
optimization equally.
The managed RAN parameter set is $\mathcal{P}=\{p_1,p_2,p_3\}$, corresponding to nominal uplink transmit power, downlink carrier bandwidth, and uplink carrier bandwidth. Each parameter has five admissible candidate levels. The KPI observation window is $K=5$ control intervals, and the SliceGuard validation window is $W=10$ intervals with rule-inconsistency threshold $\psi_{\mathrm{thresh}}=0.3$. The packet-loss degradation threshold is 1.5\%. Each epoch lasts 500 control intervals, with a traffic spike injected over intervals 200--250 by increasing the demand arrival rate of all slices by $2.5\times$. Before online control, the offline knowledge phase pre-populates the KPI database using 200 controlled configuration trials. Results are averaged over five independent runs with different random seeds.

\textbf{Baseline Methods.}
We compare SlicePlanner with six baselines: i) Static Slicing (SSlice), which keeps the initial allocation fixed; ii) Threshold-Based Slicing (TBS), which reconfigures slices after KPI-threshold violations; iii) Data-Driven Slicing (DDS), which selects the candidate with the largest predicted gain $G(v;\mathbf{x}_t)$ without logic-guided feasibility checking; iv) Deep-Reinforcement-Learning Slicing (DRL-Slicing), which uses a standalone actor-critic controller following prior A2C/PPO-based O-RAN xApp designs~\cite{kouchaki2022actor}; v) SlicePlanner without SliceGuard (SP w/o SG), which disables post-action validation and rollback while retaining pre-action feasibility checking; and vi) SlicePlanner without Learned Control Rules (SP w/o LCR), which disables $\mathcal{R}_{\mathrm{ctrl}}$ and screens candidates only using operator-specified SLA constraints.

\textbf{Simulation Results.}
\emph{\textbf{SLA Latency Violation Rate:}}
Fig.~\ref{fig:latency}(a) reports the average maximum latency SLA violation rate per slice. SSlice has the highest violations, reaching 38.2\% on eMBB and 41.5\% on URLLC, due to its fixed allocation. TBS reduces violations to 22.4\% and 28.7\% through threshold-triggered reconfiguration, while DDS reaches 16.8\% and 19.3\% using historical KPI evidence but without pre-action feasibility checking. DRL-Slicing achieves 10.5\% and 13.4\%, but remains limited by fixed parameters trained on a static distribution. The SlicePlanner ablations, SP~w/o~SG and SP~w/o~LCR, reduce violations to 8.9\%/11.2\% and 9.3\%/12.0\%, respectively. Full SlicePlanner achieves the lowest rates: 5.2\% on eMBB, 6.8\% on URLLC, and 0.3\% on mMTC. \emph{\textbf{Average Cell Throughput:}}
Fig.~\ref{fig:latency}(b) shows that SlicePlanner preserves cell throughput while improving SLA compliance. SSlice obtains 47.1~Mbps, TBS and DDS reach 48.9~Mbps and 51.3~Mbps, DRL-Slicing reaches 52.7~Mbps, and the two SlicePlanner ablations reach 53.1~Mbps and 52.8~Mbps. Full SlicePlanner achieves the highest throughput, 53.6~Mbps, indicating that feasibility checking and rollback do not reduce network utilization. \emph{\textbf{Cumulative SLA Violations:}}
Fig.~\ref{fig:latency}(c) shows cumulative SLA violations over 500 control intervals, with a traffic spike at intervals 200--250. SSlice diverges fastest before the spike, and the largest spike-induced increase is observed for SSlice and TBS. DRL-Slicing recovers more slowly than SlicePlanner under out-of-distribution traffic. SlicePlanner maintains the lowest cumulative violation count and fastest post-spike recovery, enabled by SliceGuard rollback and online KPI-memory updates. The ablations confirm that removing either SliceGuard or $\mathcal{R}_{\mathrm{ctrl}}$ weakens recovery.

\indent\underline{\emph{\textbf{SlicePlanner O-RAN Testbed Implementation.}}}
SlicePlanner is implemented with two agents: \emph{SliceSense} for decision generation and \emph{SliceGuard} for post-action validation and rollback. As shown in Fig.~\ref{fig:testbed}(a) and (b), the over-the-air testbed includes an OpenAirInterface (OAI) gNB, USRP~B210 software-defined radio, an OAI 5G Core, FlexRIC as the near-RT RIC, and two commercial Sierra Wireless 5G UEs with programmable SIM cards. The UEs are mapped to two slices through their configured Single Network Slice Selection Assistance Information (S-NSSAI) values: UE1 is assigned to a broadband slice, and UE2 is assigned to a higher-priority service slice. 
Slice reconfiguration is issued through FlexRIC over the E2 interface using RAN Control messages. The measured control-plane round-trip latency from PRB reconfiguration to the gNB control CONTROL~ACK is 390--896~\(\mu\)s. Runtime KPIs are streamed to Kinetica as online KPI memory; each record stores the UE identifier, timestamp, bitrate, packet-loss percentage, lost packets, total packets, and jitter. \emph{\textbf{SliceSense decision generation.}}
SliceSense runs with a 1~s decision period and computes per-UE loss statistics from the latest 10~s Kinetica telemetry window. A degradation event is triggered when maximum packet loss exceeds 1.5\%. If multiple UEs violate the threshold, SliceSense selects the highest-loss UE and issues the selected admissible RAN control action, increasing the degraded UE's resource share while reducing the competing allocation. \emph{\textbf{Historical telemetry and SliceGuard validation.}}
The telemetry tool retrieves recent per-UE bitrate, packet loss, packet counts, and jitter from Kinetica for candidate selection. After each reconfiguration, SliceGuard evaluates the post-action response after a 10~s settling interval and compares post-action KPIs with the pre-action baseline. The update is retained only if the degraded UE improves without harming the competing slice; otherwise, SliceGuard restores the previous values. Repeated failures update memory and trigger re-exploration when learned rules become outdated.
\textbf{Testbed Baselines.} We compare SlicePlanner with three testbed baselines. \emph{Hist. NN} selects the action associated with the nearest historical KPI state, capturing memory-based adaptation but generalizing poorly to unseen stress patterns. \emph{KPI-only XGBoost} trains a gradient-boosted decision-tree model on historical KPI-action-response records and uses numerical features, including bitrate, packet loss, jitter, and PRB utilization, to score candidates. It excludes textual reasoning, SLA constraints, validation, and rollback. \emph{LLM-only} prompts a Llama-3.1-based planner with the current KPI summary and applies the proposed action directly, without historical outcome memory, validation, or rollback. In contrast, SlicePlanner combines KPI-guided planning, operator constraints, and closed-loop post-action verification.

\textbf{Testbed Results.}
Fig.~\ref{fig:performance_load_severity_urll} evaluates robustness under increasing traffic/load severity. SlicePlanner 
introduces moderate decision latency. However, its latency remains far below 
the LLM-only baseline because the surrogate $\mathcal{M}_\theta$ produces a 
short pre-ranked action list, reducing the LLM's role to classification-like 
selection rather than open-ended action generation. This shortens the prompt 
and token output, keeping latency acceptable relative to the measured E2 
control latency. More importantly, SlicePlanner achieves the lowest packet-loss 
p95 and jitter p95 across all load levels while maintaining the highest 
throughput satisfaction. The SP~w/o~Guard ablation improves over the historical, 
KPI-only, and LLM-only baselines, but remains worse than the full system, 
confirming the value of post-action validation under high load. Fig.~\ref{fig:emd_response_fidelity_urllc_embb} reports KPI-response fidelity 
during a temporary stress interval for URLLC and eMBB slices. In the testbed, 
each control interval corresponds to a 1~s SliceSense monitoring-and-decision 
cycle that evaluates the latest 10~s KPI window from Kinetica. We use Earth 
Mover's Distance (EMD) to compare observed post-action and target KPI-response 
distributions; lower EMD indicates closer fidelity. Across bitrate, packet 
loss, jitter, and PRB utilization, SlicePlanner achieves the lowest EMD for 
both slices, showing that its measured response remains closest to the target. 
This separation is most visible during the stress and recovery periods, where 
Hist.~NN and KPI-only ML exhibit larger deviations and slower stabilization. 
LLM-only improves over historical matching in some intervals, but remains less 
stable because it lacks post-action validation and rollback. Overall, these 
results show that SlicePlanner improves KPI-response fidelity and tail QoS by 
coupling candidate selection with closed-loop verification.
\section{Conclusion}
\label{sec:conclusion}

This paper presented \emph{SlicePlanner}, a validation-aware framework
for online QoS-aware slicing in cloud-native Open RAN systems. SlicePlanner
separates KPI monitoring, candidate selection, rule-based admissibility
checking, and post-action validation into distinct control stages. By
combining historical KPI evidence with candidate-conditioned KPI-response
prediction, SlicePlanner ranks RAN configurations while screening out
SLA-infeasible or rule-inconsistent actions before they affect the live
network. SliceSense selects configurations based on predicted throughput and
latency responses, while SliceGuard validates post-action behavior, commits
beneficial changes, and rolls back harmful ones. This closed-loop design
provides an auditable and safety-aware alternative to direct model-to-action
slicing policies. \textbf{\underline{Limitations.}} First, the current controller adjusts one
parameter at a time, which improves interpretability but may miss stronger
multi-parameter configurations. Second, candidate quality depends on the
coverage and temporal relevance of the historical KPI memory; sparse data or
rapid distribution shifts can reduce reliability. Third, the learned rules are
limited to observed testbed variables and may not capture hidden RAN states or
vendor-specific scheduler behavior. Future work will extend SlicePlanner to
multi-parameter actions, calibrated uncertainty estimation, larger multi-cell
deployments, and broader evaluation under heterogeneous traffic, mobility, and
failure scenarios.

\balance
\IEEEpeerreviewmaketitle
\bibliographystyle{ieeetr}\small
\bibliography{Ref,adhwaa}

\end{document}